\documentclass[11pt,twoside]{article}
\usepackage{fancyhdr}
\usepackage[colorlinks,citecolor=blue,urlcolor=blue,linkcolor=blue,bookmarks=false]{hyperref}
\usepackage{amsfonts,epsfig,graphicx}
\usepackage{afterpage}
\usepackage{amsmath,amssymb,amsthm} 
\usepackage{fullpage}
\usepackage[T1]{fontenc} 
\usepackage{epsf} 
\usepackage{graphics} 
\usepackage{amsfonts,amsmath}
\usepackage[sort,numbers]{natbib} 
\usepackage{psfrag,xspace}
\usepackage{color,etoolbox}

\newtheorem{theorem}{Theorem} 
\newtheorem{lemma}{Lemma}

\newtheorem{definition}{Definition}

\newcommand{\vect}[1]{\ensuremath{\mathbf{#1}}}
\newcommand{\mat}[1]{\ensuremath{\mathbf{#1}}}

\newcommand{\grad}{\nabla}

\newcommand{\norm}[1]{\|{#1}\|}
\newcommand{\abs}[1]{|{#1}|}

\renewcommand{\det}{\text{det}}

\newcommand{\trans}{^{\top}}

\newcommand{\new}{^{\text{new}}}
\newcommand{\defeq}{:=}
\newcommand{\GMM}{\text{GMM}}

\newcommand{\R}{\mathbb{R}}
\newcommand{\E}{\mathbb{E}}
\newcommand{\Pb}{\mathbb{P}}

\newcommand{\gap}{C_{\mathrm{gap}}}

\newcommand{\Dmat}{\mat{D}}

\newcommand{\I}{\mathrm{I}}

\renewcommand{\v}{\vect{v}}

\newcommand{\x}{\vect{x}}

\newcommand{\g}{\vect{g}}
\newcommand{\mSigma}{\mat{\Sigma}}

\newcommand{\widgraph}[2]{\includegraphics[keepaspectratio,width=#1]{#2}}
\newcommand{\Like}{\ensuremath{\mathcal{L}}}
\newcommand{\Ball}[2]{\mathbb{B}_{#1}(#2)}
\newcommand{\Complement}[1]{\overline{#1}}

\newcommand{\xsam}{\ensuremath{x}}
\newcommand{\samind}{\ensuremath{\ell}}
\newcommand{\Xrv}{\ensuremath{X}}
\newcommand{\Event}{\ensuremath{\mathcal{E}}}
\newcommand{\Fevent}{\ensuremath{\mathcal{F}}}

\newcommand{\usedim}{\ensuremath{d}}
\newcommand{\mubold}{\ensuremath{\boldsymbol{\mu}}}
\newcommand{\lambold}{\ensuremath{\boldsymbol{\lambda}}}
\newcommand{\sep}{\ensuremath{\xi}}
\newcommand{\mixind}{\ensuremath{i}}
\newcommand{\mixtwo}{\ensuremath{j}}
\newcommand{\nummix}{\ensuremath{M}}
\newcommand{\mustar}{\ensuremath{\mu^*}}
\newcommand{\muboldstar}{\ensuremath{\mubold^*}}
\newcommand{\muboldt}{\ensuremath{\mubold^t}}

\newcommand{\numobs}{\ensuremath{n}}
\newcommand{\SamLike}{\ensuremath{\Like_\numobs}}
\newcommand{\PopLike}{\ensuremath{\Like}}
\newcommand{\Exs}{\E}
\newcommand{\thetanew}{\ensuremath{\theta^{\text{\small{new}} }}}

\newcommand{\stepsize}{\ensuremath{s}}

\newcommand{\QMAT}{\ensuremath{\mathbf{Q}}} 
\newcommand{\DMAT}{\ensuremath{\mathbf{D}}} 

\newcommand{\defn}{\ensuremath{: \, =}}
\newcommand{\muboldtilde}{\ensuremath{\tilde{\mubold}}}

\makeatletter
\long\def\@makecaption#1#2{
        \vskip 0.8ex
        \setbox\@tempboxa\hbox{\small {\bf #1:} #2}
        \parindent 1.5em  %% How can we use the global value of this???
        \dimen0=\hsize
        \advance\dimen0 by -3em
        \ifdim \wd\@tempboxa >\dimen0
                \hbox to \hsize{
                        \parindent 0em
                        \hfil 
                        \parbox{\dimen0}{\def\baselinestretch{0.96}\small
                                {\bf #1.} #2
                                } 
                        \hfil}
        \else \hbox to \hsize{\hfil \box\@tempboxa \hfil}
        \fi
        }
\makeatother

\begin{document}

\begin{center} {\Large{\bf{Local Maxima in the  Likelihood of Gaussian
  Mixture Models: Structural Results and Algorithmic Consequences}}}
\\

\vspace*{.3in}

{\large{
\begin{center}
Chi Jin$^\dagger$ ~~~~~ Yuchen Zhang$^\dagger$ ~~~~~ Sivaraman Balakrishnan$^{\ast}$ \\
\vspace{.2cm}
Martin J. Wainwright$^{\dagger,\ddagger}$ ~~~~~ Michael Jordan$^{\dagger,\ddagger}$ \\
\end{center}

\vspace*{.1in}

\begin{tabular}{c}
Department of Electrical Engineering and Computer Sciences$^{\dagger}$ \\
Department of Statistics$^{\ddagger}$ \\
\end{tabular}
\begin{tabular}{c}
University of California, Berkeley 

\vspace{.2cm}

\\ Department of Statistics$^{\ast}$ \\
Carnegie Mellon University
\end{tabular}

\vspace*{.2in}

\begin{tabular}{c}
{\texttt{chijin@cs.berkeley.edu, yuczhang@berkeley.edu, siva@stat.cmu.edu}} \\
 {\texttt{wainwrig@berkeley.edu, jordan@cs.berkeley.edu}}
\end{tabular}
}}

\vspace*{.2in}

\today
\vspace*{.2in}

\begin{abstract}
We provide two fundamental results on the population (infinite-sample)
likelihood function of Gaussian mixture models with $M \geq 3$
components. Our first main result shows that the population likelihood
function has bad local maxima even in the special case of
equally-weighted mixtures of well-separated and spherical
Gaussians. We prove that the log-likelihood value of these bad local
maxima can be arbitrarily worse than that of any global optimum,
thereby resolving an open question of~\citet{srebro2007there}. Our
second main result shows that the EM algorithm (or a first-order
variant of it) with random initialization will converge to bad
critical points with probability at least $1-e^{-\Omega(M)}$.  We
further establish that a first-order variant of EM will not converge
to strict saddle points almost surely, indicating that the poor
performance of the first-order method can be attributed to the
existence of bad local maxima rather than bad saddle points.  Overall,
our results highlight the necessity of careful initialization when
using the EM algorithm in practice, even when applied in highly
favorable settings.
\end{abstract}

\end{center}

%!TEX root = main.tex

\section{Introduction}
Finite mixture models are widely used in variety of statistical
settings, as models for heterogeneous populations, as flexible
models for multivariate density estimation and as models for 
clustering.  Their ability to model data as arising from underlying
subpopulations provides essential flexibility in a wide range
of applications~\citet{titterington1985statistical}.  This combinatorial
structure also creates challenges for statistical and computational
theory, and there are many problems associated with estimation
of finite mixtures that are still open.  These problems are often
studied in the setting of Gaussian mixture models (GMMs), reflecting
the wide use of GMMs in applications, particular in the multivariate
setting, and this setting will also be our focus in the current paper.

Early work~\citep{teicher1963identifiability} studied the
identifiability of finite mixture models, and this problem has 
continued to attract significant interest (see the recent paper of
\citet{allman2009identifiability} for a recent overview).  More
recent theoretical work has focused on issues related to the use 
of GMMs for the density estimation problem
\citep{genovese00rates,ghosal01entropies}.  Focusing on rates of
convergence for parameter estimation in GMMs, \citet{chen1995optimal}
established the surprising result that when the number of mixture
components is unknown, then the standard $\sqrt{n}$-rate for regular
parametric models is not achievable. Recent
investigations~\citep{ho2015identifiability} into exact-fitted,
under-fitted and over-fitted GMMs have characterized the achievable
rates of convergence in these settings.

From an algorithmic perspective, the dominant practical method for
estimating GMMs is the Expectation-Maximization (EM)
algorithm~\citep{dempster1977maximum}. The EM algorithm is an ascent
method for maximizing the likelihood, but is only guaranteed to converge 
to a stationary point of the likelihood function. As such, there are no
general guarantees for the quality of the estimate produced via the EM
algorithm for Gaussian mixture models.\footnote{In addition to issues
  of convergence to non-maximal stationary points, solutions of infinite 
  likelihood exist for GMMs where both the location and scale parameters 
  are estimated.  In practice, several methods exist to avoid such 
  solutions. In this paper, we avoid this issue by focusing on GMMs 
  in which the scale parameters are fixed.}
This has led researchers to explore various alternative algorithms which are
computationally efficient, and for which rigorous statistical
guarantees can be given. Broadly, these algorithms are based either on
clustering~\citep{arora2005learning,
  dasgupta2007probabilistic,vempala2002spectral,chaudhuri2008learning}
or on the method of moments~\citep{belkin2010polynomial,
  moitra2010settling,hsu2013learning}.

Although general guarantees have not yet emerged, there has nonetheless
been substantial progress on the theoretical analysis of EM and its variations.  
\citet{dasgupta2007probabilistic} analyzed a two-round variant of 
EM, which involved over-fitting the mixture and then pruning extra 
centers. They showed that this algorithm can be used to estimate Gaussian 
mixture components whose means are separated by at least $\Omega(d^{1/4})$.
\citet{balakrishnan2014statistical} studied the local convergence of
the EM algorithm for a mixture of two Gaussians with
$\Omega(1)$-separation.  Their results show that global optima
have relatively large regions of attraction, but still require that
the EM algorithm be provided with a reasonable initialization
in order to ensure convergence to a near globally optimal 
solution.

To date, computationally efficient algorithms for estimating a 
GMM provide guarantees under the strong assumption that the samples 
come from a mixture of Gaussians---i.e., that the model is well-specified. 
In practice however, we never expect the data to exactly follow the 
generative model, and it is important to understand the robustness 
of our algorithms to this assumption.  In fact, maximum likelihood
has favorable properties in this regard---maximum-likelihood estimates 
are well known to be robust to perturbations in the Kullback-Leibler 
metric of the generative model~\citep{donoho1988automatic}.  This 
mathematical result motivates further study of EM and other 
likelihood-based methods from the computational point of view.  
It would be useful to characterize when efficient algorithms can 
be used to compute a maximum likelihood estimate, or a solution 
that is nearly as accurate, and which retains the robustness
properties of the maximum likelihood estimate.

In this paper, we focus our attention on uniformly weighted mixtures
of $\nummix$ isotropic Gaussians. For this favorable setting,
\citet{srebro2007there} conjectured that any local maximum of the
likelihood function is a global maximum in the limit of infinite
samples---in other words, that there are no bad local maxima for the
population GMM likelihood function. This conjecture, if true,
would provide strong theoretical justification for EM, at least 
for large sample sizes.  For suitably small sample sizes, it is
known~\citep{amendola2015maximum} that configurations of the samples
can be constructed which lead to the likelihood function having an
unbounded number of local maxima. The conjecture
of~\citet{srebro2007there} avoids this by requiring that the samples
come from the specified GMM, as well as by considering the (infinite-sample-size) 
population setting. In the context of high-dimensional regression, 
it has been observed that in some cases despite having a non-convex 
objective function, every local optimum of the objective is within 
a small, vanishing distance of a global optimum \citep[see, e.g.,][]
{loh2013regularized,wang2014optimal}. In these settings, it is indeed 
the case that for sufficiently large sample sizes there are
no bad local optima.

\paragraph{ A mixture of two spherical Gaussians: } 
A Gaussian mixture model with a single component is simply a
Gaussian, so the conjecture of \citet{srebro2007there} holds
trivially in this case.  The first interesting case is a Gaussian mixture
with two components, for which empirical evidence supports the
conjecture that there are no bad local optima.  It is possible
to visualize the setting when there are only two components and to
develop a more detailed understanding of the population likelihood
surface.

\begin{figure}[ht]
%\vskip 0.2in
\begin{center}
\begin{tabular}{cc}
\widgraph{0.48\columnwidth}{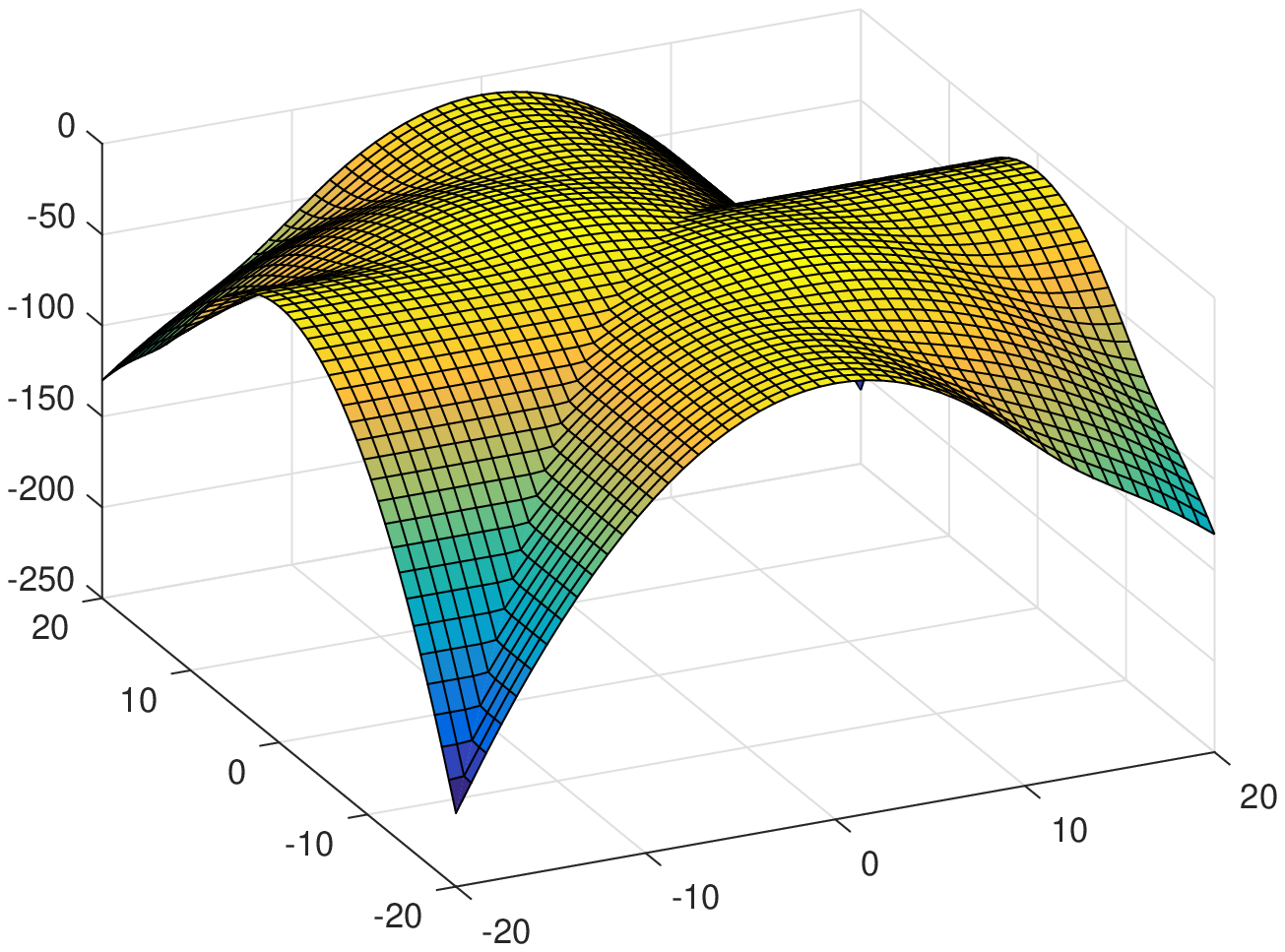} &
\widgraph{0.48\columnwidth}{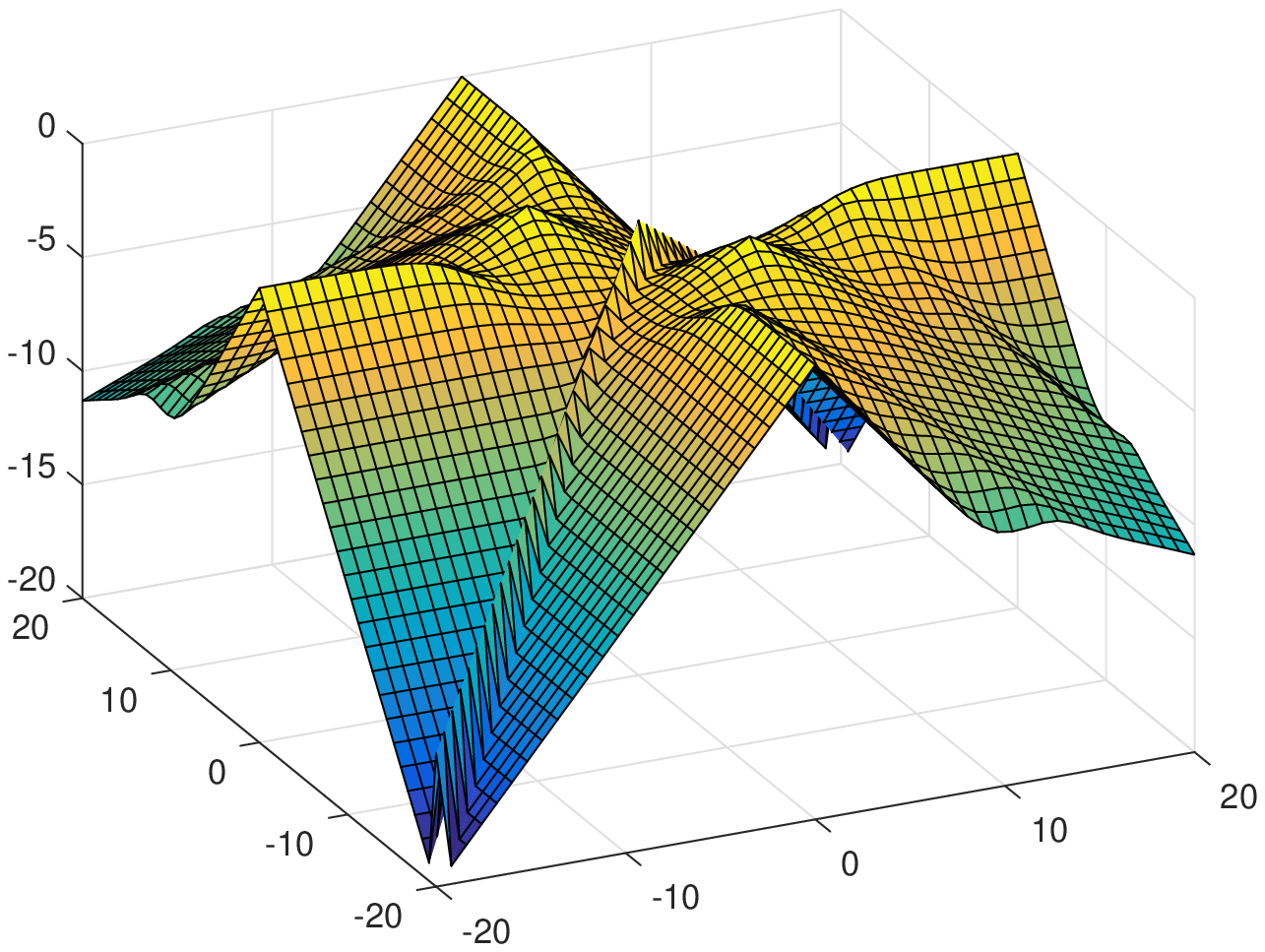} \\
(a) & (b)
\end{tabular}
\caption{Illustration of the likelihood and gradient maps for a
  two-component Gaussian mixture.  (a) Plot of population log-likehood
  map $\mubold \mapsto \Like(\mubold)$.  (b) Plot of the negative
  Euclidean norm of the gradient map $\mubold \mapsto - \|\grad
  \Like(\mubold)\|_2$. }
    %  \mjwcomment{The mesh size in these figures is much too
    % fine.  Please redraw.  Also please make MATLAB .m files that will
    % reproduce every figure in the paper.}}
\label{fig:2mixture}
\end{center}
%\vskip -0.2in
\vskip -0.1in
\end{figure}

Consider for instance a one-dimensional equally weighted unit variance
GMM with true centers $\mu_1^*=-4$ and $\mu_2^*=4$, and consider the
log-likelihood as a function of the vector $\mubold \defn (\mu_1,
\mu_2)$.  Figure~\ref{fig:2mixture} shows both the population
log-likelihood, $\mubold \mapsto \Like(\mubold)$, and the negative
2-norm of its gradient, $\mubold \mapsto - \|\grad \Like(\mubold)\|_2$.
Observe that the only local maxima are the vectors $(-4, 4)$ and $(4,
-4)$, which are both also global maxima. The only remaining critical
point is $(0, 0)$, which is a saddle point. Although points of the
form $(0, R), (R, 0)$ have small gradient when $\abs{R}$ is large, the
gradient is not exactly zero for any finite $R$. Rigorously resolving the
question of existence or non-existence of local maxima for the setting
when $\nummix = 2$ remains an open problem.

In the remainder of our paper, we focus our attention 
on the setting where there are more than two mixture components
and attempt to develop a broader understanding of likelihood 
surfaces for these models, as well as the consequences for
algorithms.

%\jccomment{The conjecture seems true supported simulation}

Our first contribution is a negative answer to the open question
of~\citet{srebro2007there}. We construct a GMM which is a uniform
mixture of three spherical unit-variance, well-separated, Gaussians
whose population log-likelihood function contains local maxima. We
further show that the log-likelihood of these local maxima can be
arbitrarily worse than that of the global maxima.  This result
immediately implies that any local search algorithm cannot exhibit
global convergence (meaning convergence to a global optimum from all
possible starting points), even on well-separated mixtures of
Gaussians.

The mere existence of bad local maxima is not a practical concern
unless it turns out that natural algorithms are frequently trapped in
these bad local maxima. 
% Our second main result shows that the EM
% algorithm, as well as a variant thereof known as the first-order EM
% algorithm, when applied to an infinite sample from a uniform,
% spherical $\nummix$-component Gaussian mixture, converges to a bad
% critical point with an exponentially high probability under certain
% natural random initialization schemes.  
Our second main result shows that the EM
algorithm, as well as a variant thereof known as the first-order EM
algorithm, with random initialization, converges to a bad
critical point with an exponentially high probability. 
In more detail, we consider
the following practical scheme for parameter estimation in an
$\nummix$-component Gaussian mixture:
\begin{enumerate}
\item[(a)] Draw $\nummix$ i.i.d.~points $\mu_1,\dots,\mu_\nummix$ 
uniformly at random from the sample set.
  % \mjwcomment{This is not a
  %   practical scheme.  How can you draw samples from an unknown
  %   distribution??  You need to clarify that this is somehow
  %   idealized..}
\item[(b)] Run the EM or first-order EM algorithm to estimate the
  model parameters, using $\mu_1,\dots,\mu_\nummix$ as the initial
  centers.
\end{enumerate}
We note that in the limit of infinite samples, the initialization 
scheme we consider is equivalent to selecting $\nummix$ initial 
centers i.i.d from the underlying mixture distribution.
We show that for a universal constant $c > 0$, with
probability at least $1-e^{-c \nummix}$, the EM and first-order EM
algorithms converge to a suboptimal critical point, whose
log-likelihood could be arbitrarily worse than that of the global
maximum.  Conversely, in order to find a solution with satisfactory
log-likelihood via this initialization scheme, one needs repeat the
above scheme exponentially many (in $\nummix$) times, and then select
the solution with highest log-likelihood.  This result strongly
indicates that repeated random initialization followed by local search
(via either EM or its first order variant) can fail to produce useful
estimates under reasonable constraints on computational complexity.

We further prove that under the same random initialization scheme, the
first-order EM algorithm with a suitable stepsize \emph{does not}
converge to a strict saddle point with probability one.  This fact
strongly suggests that the failure of local search methods for the GMM
model is due mainly to the existence of bad local optima, and not due
to the presence of (strict) saddle points.

%\jccomment{Also mention gradient EM can not converge to strict saddle
%point}

%The proof of our main result relies on the following intuition:
%suppose that the true Gaussian centers are grouped into several
%clique, so that the mutual distance within the same clique is much
%closer than the distance across different cliques. If the initial
%points $\mu_1,\dots,\mu_\nummix$ are chosen such that each clique gets
%exactly the same number of initial points as the number of its true
%centers, then running the EM algorithm will likely converge to the
%global maxima. However, the probability of this exact matching event
%is rare, and it can be shown that with high probability some clique
%will get less initial points than its true centers. If it happens,
%then we can show that no matter how many iterations the EM algorithm
%is performed, the initial points in other cliques cannot move into
%this particular clique. Thus the EM algorithm will never converge to
%the global maxima.  \jccomment{We kind of need to say each clique
%needs to has at least one center $\mu_\mixind$, otherwise, the intuition is
%misleading, also the reviewer will think our proof is trivial...This
%intuition is not exact, maybe replace it by the ones in main result?}

Our proofs introduce new techniques to reason about the structure of
the population log-likelihood, and in particular to show the existence
of bad local optima. We expect that these general ideas will aid in
developing a better understanding of the behavior of algorithms for
non-convex optimization. From a practical standpoint, our results
strongly suggest that careful initialization is required for local
search methods, even in large-sample settings, and even for extremely
well-behaved mixture models.

The remainder of this paper is organized as follows.  In
Section~\ref{sec:prelim}, we introduce GMMs, the EM algorithm, its first-order
variant and we formally set up the problem we consider.  In Section~\ref{SecMain},
we state our main theoretical results and develop some of their
implications.  Section~\ref{sec:proof} is devoted to the proofs of our
results, with some of the more technical aspects deferred to the
appendices.

%%%%%%%%%%%%%%%%%%%%%%%%%%%%%%%%%%%%%%%%%%%%%%%%%%%%%%%%%%%%%%%%%%%%%%%%%

\section{Background and Preliminaries}
\label{sec:prelim}

In this section, we formally define the Gaussian mixture model that we
study in the paper. We then describe the EM algorithm, the first-order
EM algorithm, as well as the form of random initialization that we
analyze.  Throughout the paper, we use $[\nummix]$ to denote the set
$\{1, 2, \cdots, \nummix \}$, and $\mathcal{N}(\mu, \Sigma)$ to denote
the $d$-dimensional Gaussian distribution with mean vector $\mu$ and
covariance matrix $\Sigma$.  We use $\phi( \cdot \mid \mu, \Sigma)$ to
denote the probability density function of the Gaussian distribution
with mean vector $\mu$ and covariance matrix $\Sigma$:
\begin{align}
\label{EqnGaussianDensity}
\phi(\xsam \mid \mu, \Sigma) & \defeq \frac{1}{\sqrt{(2\pi)^d
    \det(\Sigma)}} e^{-\frac{1}{2}(\xsam-\mu)\trans \Sigma^{-1}(\xsam
  - \mu)}.
\end{align}

%%%%%%%%%%%%%%%%%%%%%%%%%%%%%%%%%%%%%%%%%%%%%%%%%%%%%%%%%%%%%%%%%%%%%%%

\subsection{Gaussian Mixture Models}

A $\usedim$-dimensional Gaussian mixture model (GMM) with $\nummix$
components can be specified by a collection $\muboldstar = \{
\mustar_i, \ldots, \mustar_\nummix\}$ of $\usedim$-dimensional mean
vectors, a vector $\lambold^* = (\lambda^*_1, \ldots,
\lambda^*_\nummix)$ of non-negative mixture weights that sum to one,
and a collection $\mSigma^* = \{\Sigma^*_1, \ldots,
\Sigma^*_\nummix\}$ of covariance matrices.  Given these parameters,
the density function of a Gaussian mixture model takes the form
\begin{align*}
p(\xsam \mid \lambold^*, \muboldstar, \mSigma^*) =
  \sum_{\mixind=1}^\nummix \lambda^*_\mixind \phi(\xsam \mid
  \mustar_\mixind, \Sigma^*_\mixind),
\end{align*}
where the Gaussian density function $\phi$ was previously defined in
equation~\eqref{EqnGaussianDensity}.  In this paper, we focus on the
idealized situation in which every mixture component is equally
weighted, and the covariance of each mixture component is the
identity. This leads to a mixture model of the form
\begin{align}
  \label{EqnSimpleGMM}
  p(\xsam \mid \muboldstar) & \defeq \frac{1}{\nummix}
                              \sum_{\mixind=1}^\nummix \phi( \xsam \mid \mustar_\mixind, \I),
\end{align}
which we denote by $\GMM(\muboldstar)$.  In this case, the only
parameters to be estimated are the mean vectors
$\muboldstar = \{\mustar_\mixind\}_{\mixind=1}^\nummix$ of the
$\nummix$ components.

The difficulty of estimating a Gaussian mixture
distribution depends on the amount of separation between the mean vectors.
More precisely, for a given parameter $\sep > 0$, we say that the
$\GMM(\muboldstar)$-model is \emph{$\sep$-separated} if
\begin{align}
  \label{EqnSeparation}
  \norm{\mustar_\mixind - \mustar_\mixtwo}_2 \ge \sep, \quad \mbox{for
  all distinct pairs $\mixind$, $\mixtwo \in [\nummix]$.}
\end{align}
We say that the mixture is \emph{well-separated} if
condition~\eqref{EqnSeparation} holds for some
$\sep = \Omega(\sqrt{\usedim})$.

Suppose that we observe an i.i.d. sequence
$\{\xsam_\samind\}_{\samind=1}^\numobs$ drawn according to the
distribution $\GMM(\muboldstar)$, and our goal is to estimate the
unknown collection of mean vectors $\muboldstar$.  The sample-based
log-likelihood function $\SamLike$ is given by
\begin{subequations}
  \begin{align}
    \label{EqnDefnSamLike}
    \SamLike(\mubold) & \defeq \frac{1}{\numobs} \sum_{\samind=1}^\numobs \log
                        \Big( \frac{1}{\nummix} \sum_{\mixind=1}^\nummix \phi(\xsam_\samind \mid
                        \mu_\mixind, \I) \Big).
  \end{align}
  As the sample size $\numobs$ tends to infinity, this sample
  likelihood converges to the population log-likelihood function $\Like$
  given by
  \begin{align}
    \label{EqnDefnPopLike}
    \Like(\mubold) = \E_{\muboldstar} \log \left( \frac{1}{\nummix} \sum_{\mixind=1}^\nummix
     \phi(\Xrv \mid \mu_\mixind, \I) \right).
  \end{align}
\end{subequations}
Here $\Exs_{\muboldstar}$ denotes expectation taken over the random
vector $\Xrv$ drawn according to the model $\GMM(\muboldstar)$.

A straightforward implication of the positivity of the KL divergence
is that the population likelihood function is in fact maximized at
$\muboldstar$ (along with permutations thereof, depending on how we
index the mixture components). On the basis of empirical evidence,
\citet{srebro2007there} conjectured that this population
log-likelihood is in fact well-behaved, in the sense of having no
spurious local optima.  In Theorem~\ref{thm:main_maxima}, we show that
this intuition is false, and provide a simple example of a mixture of
$\nummix = 3$ well-separated Gaussians in dimension $\usedim = 1$,
whose population log-likelihood function has arbitrarily bad
local optima.

%%%%%%%%%%%%%%%%%%%%%%%%%%%%%%%%%%%%%%%%%%%%%%%%%%%%%%%%%%%%%%%%%%%%%%%%%

\subsection{Expectation-Maximization Algorithm}
\label{sec:em-algorithm}

A natural way to estimate the mean vectors $\muboldstar$ is by
attempting to maximize the sample log-likelihood defined by the
samples $\{x_\samind\}_{\samind=1}^\numobs$. For a non-degenerate
Gaussian mixture model, the log-likelihood is non-concave.  Rather
than attempting to maximize the log-likelihood directly, the EM
algorithm proceeds by iteratively maximizing a lower bound on the
log-likelihood.  It does so by alternating between two steps:
\begin{enumerate}
\item E-step: For each $\mixind \in [\nummix]$ and $\samind \in [n]$,
  compute the \emph{membership weight}
  \begin{align*}
    w_\mixind(\xsam_\samind) = \frac{\phi(\xsam_\samind \mid \mu_\mixind,
    \I)}{\sum_{\mixtwo = 1}^\nummix \phi(\xsam_\samind \mid \mu_\mixtwo, \I)}.
  \end{align*}
\item M-step: For each $\mixind \in [\nummix]$, update the mean
  $\mu_\mixind$ vector via
  \begin{align*}
    \mu_\mixind^{\text{new}} = \frac{\sum_{i=1}^\numobs w_\mixind(\xsam_\samind)
    \, \xsam_\samind}{\sum_{\samind=1}^\numobs w_\mixind(\xsam_\samind)}.
  \end{align*}
\end{enumerate}
In the population setting, the M-step becomes:
\begin{align}
  \label{mean_update}
  \mu_\mixind^{\text{new}} = \frac{\E_{\muboldstar} [w_\mixind(\Xrv) \,
  \Xrv]}{\E_{\muboldstar} [w_\mixind(\Xrv)]}.
\end{align}
Intuitively, the M-step updates the mean vector of each Gaussian component to be a weighted
centroid of the samples for appropriately chosen weights.
%computes a weighted centroid of the samples that are
%assigned to the $\mixind^{th}$ Gaussian component.

%%%%%%%%%%%%%%%%%%%%%%%%%%%%%%%%%%%%%%%%%%%%%%%%%%%%%%%%%%%%%%%%%%%%%%%%

\paragraph{First-order EM updates:} For a general latent variable model
with observed variables $\Xrv = x$, latent variables $Z$ and model
parameters $\theta$, by Jensen's inequality, the log-likelihood function
can be lower bounded as
\begin{align*}
  \log \Pb(\xsam \mid \theta') & \geq \underbrace{\E_{Z\sim \Pb(\cdot
                                 \mid \xsam; \theta)} \log \Pb(\xsam, Z \mid\theta')}_{\defeq
                                 Q(\theta' \mid \theta)} - \E_{Z \sim \Pb(\cdot \mid \xsam;
                                 \theta)}\log \Pb(Z \mid \xsam ;\theta').
\end{align*}
Each step of the EM algorithm can also be viewed as optimizing over
this lower bound, which gives:
\begin{align*}
  \thetanew & \defeq \arg \max_{\theta'} Q(\theta' \mid \theta)
\end{align*}
There are many variants of the EM algorithm which rely instead on
partial updates at each iteration instead of finding the exact optimum
of $Q(\theta'\mid\theta)$. One important example, analyzed in the work
of \citet{balakrishnan2014statistical}, is the first-order EM
algorithm. The first-order EM algorithm takes a step along the
gradient of the function $Q(\theta'\mid\theta)$ (with respect to its
first argument) in each iteration. Concretely, given a step size
$\stepsize > 0$, the first-order EM updates can be written as:
\begin{align*}
  \thetanew = \theta + \stepsize \grad_{\theta'} Q(\theta'\mid\theta)
  \mid_{\theta' = \theta}.
\end{align*}
In the case of the model $\GMM(\mu^*)$, the gradient EM
updates on the population objective take the form
\begin{align}
  \label{grad_EM_update}
  \mu_\mixind^{\text{new}} = \mu_\mixind + \stepsize \: \E_{\mustar}
  \big[w_\mixind(\Xrv) (\Xrv - \mu_\mixind) \big].
\end{align}
This update turns out to be equivalent to
gradient ascent on the population likelihood $\Like$ with step size
$\stepsize > 0$ (see the paper~\cite{balakrishnan2014statistical} for
details).

%%%%%%%%%%%%%%%%%%%%%%%%%%%%%%%%%%%%%%%%%%%%%%%%%%%%%%%%%%%%%%%%%%%%%%%%%

\subsection{Random Initialization}

Since the log-likelihood function is non-concave, the point to which 
the EM algorithm converges depends on the initial value 
of $\mubold$.  In
practice, it is standard to choose these values by some form of random
initialization.  For instance, one method is to to initialize the mean
vectors by sampling uniformly at random from the data set
$\{\xsam_\samind\}_{\samind=1}^\numobs$.  This scheme is intuitively
reasonable, because it automatically adapts to the locations of the
true centers. If the true centers have large mutual distances, then
the initialized centers will also be scattered. Conversely, if the
true centers concentrate in a small region of the space, then the
initialized centers will also be close to each other. In practice,
initializing $\mubold$ by uniformly drawing from the data is often more 
reasonable than drawing $\mubold$ from a fixed distribution.

In this paper, we analyze the EM algorithm and its variants at the
population level. We focus on the above practical initialization scheme of
selecting $\mubold$ uniformly at random from the sample set.
In the idealized population setting, this is equivalent to sampling the initial
values of $\mubold$ i.i.d from the distribution
$\GMM(\muboldstar)$. Throughout this paper, we refer to this particular
initialization strategy as \emph{random initialization.}

% \mjwcomment{Again somehow need to clarify that this is an idealized
% form of random initialization...I don't know how you can sample from
% the unknown distribution...}

% !TEX root = main.tex

%%%%%%%%%%%%%%%%%%%%%%%%%%%%%%%%%%%%%%%%%%%%%%%%%%%%%%%%%%%%%%%%%%%%%%%%%%

\section{Main results} 
\label{SecMain}

We now turn to the statements of our main results, along with a
discussion of some of their consequences.

%%%%%%%%%%%%%%%%%%%%%%%%%%%%%%%%%%%%%%%%%%%%%%%%%%%%%%%%%%%%%%%%%%%%%

\subsection{Structural properties}

In our first main result (Theorem~\ref{thm:main_maxima}), for any
$\nummix \geq 3 $, we exhibit an $\nummix$-component mixture of
Gaussians in dimension $d=1$ for which the population log-likelihood
has a bad local maximum whose log-likelihood is arbitrarily worse than
that attained by the true parameters $\muboldstar$. This result
provides a negative answer to the conjecture
of~\citet{srebro2007there}.

% \mjwcomment{Please note how this theorem has been rewritten to make it
%   shorter, ``punchier'' and clearer.  Please do the same for the other
%   theorems.  Avoid unnecessary clutter in theorem statements.}

\begin{theorem}
\label{thm:main_maxima} For any $\nummix \ge 3$ and any constant
$\gap>0$, there is a well-separated uniform mixture of $\nummix$
unit-variance spherical Gaussians $\GMM(\muboldstar)$ and a local
maximum $\mubold'$ such that
\begin{equation*}
\Like(\mubold') \leq \Like(\muboldstar) - \gap.
\end{equation*}
\end{theorem}
In order to illustrate the intuition underlying
Theorem~\ref{thm:main_maxima}, we give a geometrical description of our
construction for $\nummix = 3$. Suppose that the true centers
$\mu_1^*$, $\mu_2^*$ and $\mu_3^*$, are such that the distance between
$\mu_1^*$ and $\mu_2^*$ is much smaller than the respective distances
from $\mu_1^*$ to $\mu_3^*$, and from $\mu_2^*$ to $\mu_3^*$.  Now,
consider the point $\mubold \defeq (\mu_1,\mu_2,\mu_3)$ where $\mu_1 =
(\mu_1^* + \mu_2^*)/2$; the points $\mu_2$ and $\mu_3$ are both placed
at the true center $\mu_3^*$.  This assignment does not maximize the
population log-likelihood, because only one center is assigned to the
two Gaussian components centered at $\mu_1^*$ and $\mu_2^*$, while two
centers are assigned to the Gaussian component centered at $\mu_3^*$.
However, when the components are well-separated we are able to show
that there is a local maximum in the neighborhood of this
configuration.  In order to establish the existence of a local
maximum, we first define a neighborhood of this configuration ensuring
that it does not contain any global maximum, and then prove that the
log-likelihood on the boundary of this neighborhood is strictly smaller
than that of the sub-optimal configuration $\mubold$.  Since the
log-likelihood is bounded from above, this neighborhood must contain
at least one maximum of the log-likelihood. Since the global maxima
are not in this neighborhood by construction, any maximum in this
neighborhood must be a local maximum. See Section~\ref{sec:proof} for
a detailed proof.

%%%%%%%%%%%%%%%%%%%%%%%%%%%%%%%%%%%%%%%%%%%%%%%%%%%%%%%%%%%%%%%%%%%%%%%%%

\subsection{Algorithmic consequences}

An important implication of Theorem~\ref{thm:main_maxima} is that any
iterative algorithm, such as EM or gradient ascent, that attempts to
maximize the likelihood based on local updates \emph{cannot} be
globally convergent---that is, cannot converge to (near) globally
optimal solutions from an arbitrary initialization. Indeed, if any
such algorithm is initialized at the local maximum, then they will
remain trapped.  However, one might argue that this conclusion
is overly pessimistic, in that we have only shown that these
algorithms fail when initialized at a certain (adversarially chosen)
point.  Indeed, the mere existence of bad local minima need not be a
practical concern unless it can be shown that a typical optimization
algorithm will frequently converge to one of them.  The following
result shows that the EM algorithm, when applied to the population
likelihood and initialized according to the random scheme described in
Section~\ref{sec:em-algorithm}, converges to a bad critical point with
high probability.

\begin{theorem}
\label{thm:main_EM}
Let $\muboldt$ be the $t^{\mathrm{th}}$ iterate of the EM algorithm
initialized by the random initialization scheme described previously.
There exists a universal constant $c$, for any $\nummix \geq 3$ and 
any constant $\gap>0$, such that there is a well-separated 
uniform mixture of $\nummix$ unit-variance spherical
Gaussians $\GMM(\muboldstar)$ with
\begin{align*}
\Pb \left[ \forall t \ge 0, ~ \Like(\muboldt) \leq \Like(\muboldstar)
  - \gap \right] \ge 1 - e^{-c \nummix}.
\end{align*}
\end{theorem}
% \mjwcomment{Logic clarification:  does the ``universal'' constant depend
% on $\nummix$ and/or $\gap$?  If not, then you should write: ``There is
% a universal constant $c$ such that for $\nummix \geq 3$.....''}

Theorem~\ref{thm:main_EM} shows that, for the specified configuration
$\muboldstar$, the probability of success for the EM algorithm is
exponentially small as a function of $\nummix$.  As a consequence, in
order to guarantee recovering a global maximum with at least constant
probability, the EM algorithm with random initialization must be
executed at least $e^{\Omega(\nummix)}$ times. This result strongly
suggests that that effective initialization schemes, such as those 
based on pilot estimators utilizing the method of moments~\citep{moitra2010settling,
hsu2013learning}, are critical to finding good maxima in general GMMs.

The key idea in the proof of Theorem~\ref{thm:main_EM} is the
following: suppose that all the true centers are grouped into two
clusters that are extremely far apart, and suppose further that we
initialize all the centers in the neighborhood of these two clusters,
while ensuring that at least one center lies within each cluster.  In
this situation, all centers will remain trapped within the cluster in
which they were first initialized, irrespective of how many steps we
take in the EM algorithm.  Intuitively, this suggests that the only
favorable initialization schemes (from which convergence to a global
maximum is possible) are those in which (1) all initialized centers
fall in the neighborhood of exactly one cluster of true centers, (2) the
number of centers initialized within each cluster of true centers
exactly matches the number of true centers in that cluster.  However,
this observation alone only suffices to guarantee that the success
probability is polynomially small in $\nummix$.

In order to demonstrate that the success probability is
\emph{exponentially small} in $\nummix$, we need to further refine
this construction.  In more detail, we construct a Gaussian mixture
distribution with a recursive structure: on top level, its true
centers can be grouped into two clusters far apart, and then inside
each cluster, the true centers can be further grouped into two
mini-clusters which are well-separated, and so on.  We can repeat this
structure for $\Omega(\log\nummix)$ levels.  For this GMM instance,
even in the case where the number of true centers exactly matches the
number of initialized centers in each cluster at the top level, we
still need to consider the configuration of the initial centers within
the mini-clusters, which further reduces the probability of success
for a random initialization.  A straightforward calculation then shows
that the probability of a favorable random initialization is 
on the order of $e^{-\Omega(\nummix)}$.  The full proof is
given in Section~\ref{sec:proofEM}.

We devote the remainder of this section to a treatment of the
first-order EM algorithm.  Our first result in this direction shows
that the problem of convergence to sub-optimal fixed points remains a
problem for the first-order EM algorithm, provided the step-size is
not chosen too aggressively.

\begin{theorem}\label{thm:main_grad_EM}
Let $\muboldt$ be the $t^{\mathrm{th}}$ iterate of the first-order EM
algorithm with stepsize $s \in (0,1)$, initialized by the random
initialization scheme described previously.  There exists a universal constant $c$, 
for any $\nummix \geq 3$ and any constant $\gap>0$, 
such that there is a well-separated uniform mixture of
$\nummix$ unit-variance spherical Gaussians $\GMM(\muboldstar)$ with
\begin{align}
\Pb \left( \forall t \ge 0, ~ \Like(\muboldt) \leq \Like(\muboldstar)
- \gap \right) \ge 1 - e^{-c \nummix}.
\end{align}
\end{theorem}
% \mjwcomment{Same concern here regarding the universal constant:
%   current writing seems to suggest that it may depend on $(\nummix,
%   \gap)$.}  
We note that the restriction on the step-size is weak, 
and is satisfied by the theoretically optimal choice for a
mixture of two Gaussians in the setting studied
by~\citet{balakrishnan2014statistical}.  Recall that the first-order
EM updates are identical to gradient ascent updates on the
log-likelihood function. As a consequence, we can conclude that the
most natural local search heuristics for maximizing the log-likelihood
(EM and gradient ascent), fail to provide statistically meaningful
estimates when initialized randomly, unless we repeat this procedure
exponentially many (in $\nummix$) times.

%when initialized randomly with a reasonable number of random restarts. 
%Recall for uniform weighted spherical Gaussian mixture models,
%gradient EM updates is identical to gradient ascent updates. 
%This implies Theorem \ref{thm:main_EM} also holds for gradient ascent on the 
%log-likelihood function. 

%
%Therefore, we can now conclude that for GMM, most popular
%local search methods can fail to provide statistical meaningful
%estimate even with reasonable number of repeated random
%initializations.

Our final result concerns the type of fixed points reached by the
first-order EM algorithm in our setting.  \citet{pascanu2014saddle}
argue that for high-dimensional optimization problems, the principal
difficulty is the proliferation of saddle points, not the existence 
of poor local maxima. In our setting, however, we can leverage recent 
results on gradient methods~\citep{lee2016gradient,panageas2016gradient} 
to show that the first-order EM algorithm cannot converge to strict 
saddle points.  More precisely:
\begin{definition}[Strict saddle point~\cite{ge2015escaping}]
\label{DefnStrict}
For a maximization problem, we say that a critical point $\x_{ss}$ of
function $f$ is a \emph{strict saddle point} if the Hessian $\grad^2
f(\x_{ss})$ has at least one strictly positive eigenvalue.
\end{definition}

\noindent With this definition, we have the following:
\begin{theorem}
\label{prop:escape_saddle}
Let $\muboldt$ be the $t^{\mathrm{th}}$ iterate of the first-order EM
algorithm with constant stepsize $s \in (0, 1)$, and initialized by
the random initialization scheme described previously. Then for any
$\nummix$-component mixture of spherical Gaussians:
\begin{enumerate}
\item[(a)] The iterates $\muboldt$ converge to a critical point of the
  log-likelihood.
\item[(b)] For any strict saddle point $\boldsymbol{ \mu}_{ss}$, we
  have $\mathbb{P} \left( \lim_{t \rightarrow \infty} \muboldt =
  {\boldsymbol{ \mu}_{ss} } \right) = 0$.
\end{enumerate}
\end{theorem}

\noindent Theorems~\ref{thm:main_grad_EM} and~\ref{prop:escape_saddle}
provide strong support for the claim that the sub-optimal points to which
the first-order EM algorithm frequently converges are bad local
maxima. The algorithmic failure of the first-order EM algorithm is
most likely due to the presence of bad local maxima, as opposed to
(strict) saddle-points.

The proof of Theorem~\ref{prop:escape_saddle} is based on
recent work~\citep{lee2016gradient,panageas2016gradient} on the
asymptotic performance of gradient methods.  That work reposes 
on the stable manifold theorem from dynamical systems theory, 
and, applied directly to our setting, would require establishing 
that the population likelihood $\Like$ is smooth.  Our proof
technique avoids such a smoothness argument; see Section~\ref{sec:app2} 
for the details.  The proof technique makes use of specific properties 
of the first-order EM algorithm that do not hold for the EM algorithm.
We conjecture that a similar result is true for the EM algorithm; 
however, we suspect that a generalized version of the stable manifold 
theorem will be needed to establish such a result. 

%%%%%%%%%%%%%%%%%%%%%%%%%%%%%%%%%%%%%%%%%%%%%%%%%%%%%%%%%%%%%%%%%%%%%%%%%%%%

\section{Proofs}
\label{sec:proof}

This section is devoted to the proofs of
Theorems~\ref{thm:main_maxima} through~\ref{prop:escape_saddle}.
Certain technical aspects of the proofs are deferred to the appendix.

%%%%%%%%%%%%%%%%%%%%%%%%%%%%%%%%%%%%%%%%%%%%%%%%%%%%%%%%%%%%%%%%%%%%%%

\subsection{Proof of Theorem~\ref{thm:main_maxima}} 

In this section, we prove Theorem \ref{thm:main_maxima}.  The proof
consists of three parts: starting with the case $\nummix = 3$, the
first part shows the existence of a local maximum for certain GMMs,
whereas the second part shows that this local maximum has
a log-likelihood that is much worse than that of the global maximum.
The third part provides the extension to the general case of $\nummix
> 3$ mixture components.

%%%%%%%%%%%%%%%%%%%%%%%%%%%%%%%%%%%%%%%%%%%%%%%%%%%%%%%%%%%%%%%%%%

\subsubsection{Existence of a local maximum} 

In this section, we prove the existence of a local maximum by first
constructing a family of GMMs parametrized by a scalar $\gamma$, and
then proving the existence of local maxima in the limiting case when
$\gamma \rightarrow +\infty$. By continuity of the log-likelihood
function, we can then conclude that there exists some finite $\gamma$
whose corresponding log-likelihood has local maxima.

We begin by considering the special case of $\nummix=3$ components in
dimension $d=1$. For parameters $R > 0$ and $\gamma \gg 1$, suppose
that the true centers $\muboldstar$ are given by
\begin{align*}
\mu_1^* = -R, \quad\quad \mu_2^* = R, \quad\quad \mu_3^* = \gamma R.
\end{align*}
By construction, the two centers $\mu_1^*$ and $\mu_2^*$ are
relatively close together near the origin, while the third center
$\mu_3^*$ is located far away from both of the first two centers.

We first claim that when $\gamma$ is sufficiently large, there is a
local maximum in the closed set:
\begin{align*}
\mathcal{D} = \left\{(\mu_1,\mu_2,\mu_3) \in \R^3 \, \mid \, ~\mu_1
\leq \frac{\gamma R}{3},~\mu_2\geq \frac{2\gamma R}{3}~\mbox{ and
}~\mu_3\geq \frac{2\gamma R}{3}\right\}.
\end{align*}
To establish this claim, we consider the value of population
log-likelihood function $\Like(\muboldtilde)$ at an interior point
$\tilde{\mubold} = (0, \gamma R, \gamma R)$ of $\mathcal{D}$, and
compare it to the log-likelihood on the boundary of the set
$\mathcal{D}$. We show that for a sufficiently large $\gamma$, the
log-likelihood at the interior point is strictly larger than the
log-likelihood on the boundary, and use this to argue that there must
be a local maxima in the set $\mathcal{D}$. Concretely, define $v_0
\defn \Like(\tilde{\mubold})$, and the maximum value of
$\Like(\mubold)$ on the three two-dimensional faces of $\mathcal{D}$,
i.e.,
\begin{align*}
v_1 \defn \sup_{\substack{\mu_1 = \gamma R/3\\ \mu_2 \geq 2\gamma R/3
    \\
\mu_3 \geq 2\gamma R/3}} \Like(\mubold),\quad\quad v_2 \defn
\sup_{\substack{\mu_1 \leq \gamma R/3\\ \mu_2 = 2\gamma R/3\\ \mu_3
    \geq 2\gamma R/3}} \Like(\mubold),\qquad \mbox{and} \qquad v_3
\defn \sup_{\substack{\mu_1 \leq \gamma R/3\\ \mu_2 \geq 2\gamma
    R/3\\ \mu_3 = 2\gamma R/3}} \Like(\mubold).
\end{align*}
The population log-likelihood function is given by the expression
\begin{align*}
\Like(\mubold) = \E_{\muboldstar} \log\left( \sum_{\mixind=1}^3
e^{-\frac{1}{2}(X - \mu_\mixind)^2} \right)- \log(3\sqrt{2\pi}).
\end{align*}
As $\gamma\to\infty$, it is easy to verify that 
\begin{align*}
v_0 =
\Like(\tilde{\mubold}) \to -\frac{2R^2+3 -2\log(2)}{6} -
\log(3\sqrt{2\pi}).
\end{align*} 
Similarly, we can calculate the value of $v_1,v_2$ and $v_3$ as 
$\gamma \rightarrow \infty$; i.e., a straightforward calculation 
shows that
\begin{align*}
&\lim_{\gamma\to+\infty} v_1 = -\infty,\\ &\lim_{\gamma\to+\infty} v_2
  = -\frac{2R^2+3}{6} - \log(3\sqrt{2\pi})\quad \mbox{(the maximum is
    attained at $\mu_1\to 0$ and $\mu_3\to \gamma
    R$)},\\ &\lim_{\gamma\to+\infty} v_3 = -\frac{2R^2+3}{6} -
  \log(3\sqrt{2\pi})\quad \mbox{(the maximum is attained at $\mu_1\to
    0$ and $\mu_2\to \gamma R$)}.
\end{align*}
This gives the relation $v_0 > \max\{v_1,v_2,v_3\}$ when
$\gamma\to\infty$.  Since $\Like$ is a continuous function of
$\gamma$, we know that $v_0, v_1, v_2, v_3$ are also continuous
functions of $\gamma$. Therefore, there exists a finite $A$ such that,
as long as $\gamma>A$, we will still have $v_0 > \max\{v_1,v_2,v_3\}$.
This in turn implies that the function value at an interior point is
strictly greater than the function value on the boundary of
$\mathcal{D}$, which implies the existence of at least one local
maximum inside $\mathcal{D}$.

On the other hand, the global maxima of the population
likelihood function are $(-R, R, \gamma R)$ and its permutations, which
are not in $\mathcal{D}$. This shows the existence of at
least one local maximum which is not a global maximum.

%%%%%%%%%%%%%%%%%%%%%%%%%%%%%%%%%%%%%%%%%%%%%%%%%%%%%%%%%%%%%%%%

\subsubsection{Log-likelihood at a local maximum} 

In order to prove that the log-likelihood of a local maximum can be
arbitrarily worse than the log-likelihood of the global maximum, we
consider the limit when $R \to \infty$. In this case, the limiting
value of the global maximum will be
\begin{align*}
\lim_{R \rightarrow \infty} \Like(\muboldstar)& = -\frac{1}{2} -
\log(3\sqrt{2\pi}).
\end{align*}
Let $\mubold' = (\mu'_1, \mu'_2, \mu'_3)$ be one of the local maxima
in the closed set $\mathcal{D}$. We have previously established the
existence of such a local maximum.

Since $\mu_1^* - \mu_2^* = 2R$, we know that either $\mid\mu_2^* -
\mu_1'\mid > R$ or $\mid\mu_1^* - \mu_1'\mid > R$ has to be true.
Without loss of generality, we may assume that $\mid\mu_2^* -
\mu_1'\mid > R$. From the definition of the set $\mathcal{D}$, we can
also see that $\mid\mu_2^* - \mu_2'\mid > R$ and $\mid\mu_2^* -
\mu_3'\mid > R$. Putting together the pieces yields
\begin{align*}
\lim_{R \to +\infty} \Like(\mubold') \le \lim_{R \to +\infty}
\frac{1}{3}\E_{X\sim \mathcal{N}(\mu_2^*, 1)}
\log\left(\sum_{\mixind=1}^3 e^{-\frac{1}{2}(X - \mu'_\mixind)^2}
\right)- \frac{1}{3}\log(3\sqrt{2\pi}) =-\infty.
\end{align*}
Again, by the continuity of the function $\Like$ with respect to $R$,
we know for any $\gap >0$, there always exists a large constant $A'$,
so that if $R>A'$, we will have $\Like(\muboldstar) - \Like(\mubold')
>\gap$.  This completes the proof for case $\nummix = 3$.

\subsubsection{Extension to the case $\nummix > 3$ } 

We now provide an outline of how this argument can be extended to the
general setting of $\nummix > 3$.  Consider a GMM with true centers
\begin{align*}
\mu_\mixind^* = \frac{(2\mixind-k)R}{k-2}, \qquad \mbox{for $\mixind =
  1, \cdots, \nummix-1$} \quad \mbox{and} \qquad \mu_\nummix^* =
\gamma R,
\end{align*}
for some parameter $\gamma > 0$ to be chosen.  We claim that when
$\gamma$ is sufficiently large, there is at least one local maximum in
the closed set
\begin{align*}
\mathcal{D}_\nummix = \left\{(\mu_1, \cdots, \mu_\nummix) \: \mid \;
\mu_1 \leq \frac{\gamma R}{3},~\mu_2\geq \frac{2\gamma R}{3}, \cdots,
\mu_\nummix\geq \frac{2\gamma R}{3}\right\}.
\end{align*}
The proof follows from an identical argument as in the $\nummix=3$ case. 

%%%%%%%%%%%%%%%%%%%%%%%%%%%%%%%%%%%%%%%%%%%%%%%%%%%%%%%%%%%%%%%%%%%%%%%%%%%

\subsection{Proof of Theorem~\ref{thm:main_EM}}
\label{sec:proofEM}

In this section, we prove Theorem \ref{thm:main_EM}.  We first present
an important technical lemma that addresses the behavior of the EM
algorithm for a particular configuration of true and initial centers.
We then prove the theorem by constructing a bad example and
recursively applying this lemma.  The proof of this lemma is given in
Appendix~\ref{sec:app}.

We focus on the one-dimensional setting throughout this proof.  We use
$\Ball{x}{\delta}$ to denote an interval centered at $x$ with radius
$\delta$, that is, $\Ball{x}{\delta} =[x-\delta, x+\delta]$. We also
use $\Complement{\Ball{x}{\delta}}$ to represent the complement of the
interval $\Ball{x}{\delta}$, i.e. $\Complement{\Ball{x}{\delta}} =
(-\infty,x - \delta) \cup (x + \delta, \infty).$

%We first present a lemma which tells very
%important property of EM algorithm.

% \mjwcomment{This is a poorly stated lemma: too wordy and yet unclear
%   at the same time.  A mathematical statement needs to be ``set up''
%   properly, so that it comes through cleanly.  This means that you
%   should introduce some of the assumptions prior to the lemma, so that
%   the lemma is short (at most two to three shortish sentences.}

As a preliminary, let us define a class of GMMs, which we refer to as
\emph{diffuse GMMs}.  We say that a mixture model $\GMM(\muboldstar)$
consisting of $\widetilde{\nummix}$ components is
$(c,\delta)$-\emph{diffuse} if:
\begin{enumerate}
\item[(a)] For some $\nummix \leq \widetilde{\nummix}$, there are
  $\nummix$ centers contained in $\Ball{c\delta}{\delta} \cup
  \Ball{-c\delta}{\delta}$;
\item[(b)] Each of the sets $\Ball{c\delta}{\delta}$ and
  $\Ball{-c\delta}{\delta}$ contain at least one center;
\item[(c)] The remaining $\widetilde{\nummix} - \nummix$ centers are
  all in $\Complement{\Ball{0}{20c\delta}}$.
\end{enumerate}

\noindent Consider the EM algorithm, and denote by $\nummix^{(t)}_1,
\nummix^{(t)}_2$ and $\nummix^{(t)}_3$ the number of centers the EM
algorithm has in the t$^\mathrm{th}$ iteration in the sets
$\Ball{-c\delta}{2\delta}, \Ball{c\delta}{2\delta}$ and
$\Complement{\Ball{0}{20c\delta}}$ respectively, where $c$ and
$\delta$ are those specified in the definition of the diffuse GMM. To
be clear, $\nummix^{(0)}_1, \nummix^{(0)}_2$ and $\nummix^{(0)}_3$
denote the number of centers in these sets in the initial
configuration specified to the EM algorithm. With these definitions in
place, we can state our lemma.

\begin{lemma}
\label{lem:case_strong} 
Suppose that the true underlying distribution is a
$(c,\delta)$-diffuse GMM
with $c > 20$ and $\delta > \log M + 3$, and that the EM algorithm is 
initialized so that $\nummix^{(0)}_1, \nummix^{(0)}_2 \geq 1$.
\begin{enumerate}
\item[(a)] If $M = \widetilde{M}$, then 
\begin{align}
\label{EqnMammoth}
  \nummix^{(t)}_1 = \nummix^{(0)}_1 \quad \mbox{and} \quad
  \nummix^{(t)}_2 = \nummix^{(0)}_2 \qquad \mbox{ for every $t \geq
    0$.}
\end{align}

\item[(b)] If $M < \widetilde{M}$, suppose further that for each center in $\mu_j^* \in
  \Complement{\Ball{0}{20c\delta}}$, there is an initial center
  $\mu^{(0)}_{\mixtwo'}$ such that
  \mbox{$\abs{\mu^{(0)}_{\mixtwo'}-\mu_\mixtwo^*} \le
    \abs{\mu_\mixtwo^*}/10$.} Then the same
  conditions~\eqref{EqnMammoth} hold.
\end{enumerate}
\end{lemma} 

\noindent Intuitively, these results show that if the true centers are
clustered together into two clusters that are well separated, and the
EM algorithm is initialized so that each cluster is accounted for by
at least one initial center then the EM algorithm remains trapped in
the initial configuration of centers. A concrete implication of part
(a) is that if the true distribution is a $(c,\delta)$-diffuse GMM
with $\widetilde{M} = M$ and $\nummix_1^*, \nummix_2^*$ true clusters
lie in $\Ball{-c\delta}{\delta}$ and $\Ball{c\delta}{\delta}$
respectively, then there are only three possible ways to initialize
the EM algorithm that might possibly converge to a global maximum of
the log-likelihood function; i.e., the pair $(\nummix^{(0)}_1,
\nummix^{(0)}_2)$ must be one of $\{(\nummix,0), (\nummix_1^*,
\nummix_2^*), (0,\nummix)\}$, where $\nummix = \nummix_1^* +
\nummix_2^*$.

We are now equipped to prove Theorem~\ref{thm:main_EM}.  We will first
focus on the case $\nummix=2^m$ for some positive integer $m$; the
case of arbitrary $\nummix$ will be addressed later.  At a high level,
we will first construct the distribution $\GMM(\muboldstar)$ that
establishes the theorem, and then use the above technical lemma in order
to reason about the behavior of the EM algorithm on this distribution.

\paragraph{Case $\nummix=2^m$:}

First, define the collection of $2^m$ binary vectors of the form
$\boldsymbol{\epsilon} = (\epsilon_1, \epsilon_2, \cdots, \epsilon_m)$
where each $\epsilon_i \in \{-1,1\}.$ Consider the distribution,
$\GMM(\muboldstar)$, with the locations of the true centers indexed by
these $2^m$ vectors; i.e., each center is located at
\begin{align} \label{hard_problem}
\mu(\boldsymbol{\epsilon}) = \sum_{\mixind=1}^m \epsilon_\mixind \Big(\frac{1}{100}\Big)^{\mixind-1} R,
\end{align}
where we choose $R \geq 100^{m+1} (\nummix+1)$. This in turn implies
that the distance between the closest pair of true centers is at least
$10^4\times (\nummix+1)$.

Our random initialization strategy samples the initial centers $\mu_1,
\mu_2, \cdots, \mu_\nummix$ i.i.d.~from the distribution
$\GMM(\muboldstar)$. We can view this sampling process as two separate
steps:
\begin{enumerate}
\item[(i)] Sample an integer $Z_\mixind$ uniformly from $[\nummix]$.
\item[(ii)] Sample value $\mu_\mixind$ from the Gaussian distribution
  $\mathcal{N}(\mu^*_{Z_\mixind}, \I)$.
\end{enumerate}
Concentration properties of the Gaussian distribution will ensure that
$\mu_\mixind$ will not be too far from its expectation
$\mu^*_{Z_\mixind}$. Formally, we define the following event:
\begin{align}
\label{eqn:eventnummix}
\Event_\nummix & \defeq \Big \{ \text{all~} \nummix \text{~initial
  points~} \mu_\mixind \text{~are contained
  in~}\Ball{\mu^*_{Z_\mixind}}{\nummix} \Big \}.
\end{align} 
By standard Gaussian tail bounds, we have
\begin{align*}
\Pb(\Event_\nummix)= (1-\Pb_{X\sim\mathcal{N}(0, 1)}(\abs{X} >
\nummix))^\nummix \ge (1 - 2 M e^{-\nummix^2/2} ) \ge 1 -
e^{-\Omega(\nummix)}
\end{align*}
This implies that the event $\Event_\nummix$ will hold with high
probability (when $\nummix$ is large).  Conditioned on the event
$\Event_\nummix$, we are guaranteed that all initialized points are
relatively close to some true center.

A key observation regarding the configuration of centers in the model
$\GMM(\muboldstar)$ specified by equation~\eqref{hard_problem} is that
the true centers can be partitioned into two well separated
regions. More precisely, it is easy to verify that there are
$\nummix/2$ true centers in the interval $\Ball{-R}{R/99}$ while the
remaining $\nummix/2$ true centers are contained in the interval
$\Ball{R}{R/99}$. In what follows, we refer to $\Ball{-R}{2R/99}$ as
the \emph{left urn} and to $\Ball{R}{2R/99}$ as the \emph{right urn}.
%We define \emph{left urn} as a shorthand
%notation for $\Ball{-R}{2R/99}$ as a slightly larger region that
%contains the first interval and the \emph{right urn} as a shorthand
%notation for $\Ball{R}{2R/99}$ as a slightly larger region that
%contains the second interval.

Conditioned on $\Event_\nummix$, each initial point lands in either
the left urn or the right urn with equal probability. Suppose we
initialize EM with $(\nummix_1, \nummix_2)$ centers in the left and
right urn respectively. By Lemma~\ref{lem:case_strong}(a), the only
three possible values of the pair $(\nummix_1, \nummix_2)$ for which
the EM algorithm might converge to a global optimum are $(0, \nummix),
(\nummix, 0), (\nummix/2, \nummix/2)$. A simple calculation will show
that the first and second possibilities occur with exponentially small
probability. However, the third possibility occurs with only
polynomially small probability, and so we need to further investigate
this possibility.

Consider, for example, the left urn: the true centers in the left urn
can further be partitioned into two intervals $\Ball{-1.01R}{R/9900}$
and $\Ball{-0.99R}{R/9900}$ with $\nummix/4$ true centers in
each. Thus, each urn can be further partitioned into a left urn and a
right urn. Following the same analysis as above and now using part (b)
of Lemma~\ref{lem:case_strong} instead of part (a), we see that in
order to ensure that the EM algorithm converges to a global optimum,
the number of initial centers in $\Ball{-1.01R}{2R/9900}$ and
$\Ball{-0.99R}{2R/9900}$ must be one of the following pairs $\{(0,
\nummix/2),(\nummix/2, 0), (\nummix/4, \nummix/4)\}.$

Our configuration of centers in equation~\eqref{hard_problem}
guarantees that this argument can be recursively applied until we
reach an interval which contains only two true centers.
%Given $\nummix$ true centers constructed by
%equation~\eqref{hard_problem} and 
For a configuration of $\nummix$ initial centers, we call these
initial centers a \emph{good initialization} for a collection of true 
centers if one of 
the following holds:
\begin{enumerate}
\item[(a)] $\nummix = 1$,
\item[(b)] the number of initial centers assigned to the left urn and
  the right urn of the collection of true centers
  are either $(0,\nummix)$ or $(\nummix,0)$,
\item[(c)] the number of initial centers assigned to the left urn and
  the right urn of the collection of true centers
 are $(\nummix/2,\nummix/2)$; and further recursively
  the initialization in both the
  left and the right urns are good initializations.
\end{enumerate}
Lemma~\ref{lem:case_strong} implies that the EM algorithm converges to
a global maximum only if a good initialization is realized. 
We will now 
show that the probability of a good initialization is
exponentially small.

Let $\Fevent_\nummix$ represent the event that a good initialization
is generated on a mixture with $\nummix$ components, for our
configuration of true centers.  Let $\nummix_1$ and $\nummix_2$
represent the number of initial centers in the left urn and the right
urn, respectively. Conditioning on the event $\Event_\nummix$ from
equation~\eqref{eqn:eventnummix}, we have
\begin{align*}
\Pb(\Fevent_\nummix \mid \Event_\nummix) & \le \Pb(\nummix_1=0) +
\Pb(\nummix_1 = \nummix) + \Pb(\nummix_1 = \nummix/2)\cdot
\left(\Pb(\Fevent_{\nummix/2} \mid \Event_\nummix)\right)^2 \\
& \le 2 \times \binom{\nummix}{0} \frac{1}{2^\nummix} +
\binom{\nummix}{\nummix/2} \frac{1}{2^\nummix}
\cdot\left(\Pb(\Fevent_{\nummix/2} \mid \Event_\nummix)\right)^2 \\
& \le \frac{1}{2^{\nummix-1}} + \frac{1}{2}\cdot\left(
\Pb(\Fevent_{\nummix/2} \mid \Event_\nummix)\right)^2.
\end{align*}
Since $\Pb(\Fevent_1 \mid \Event_\nummix) = 1$, solving this recursive
inequality implies that 
$\Pb(\Fevent_\nummix \mid \Event_\nummix) \le e^{-c
  \nummix}$ for some universal constant $c$. 
  %We know that if event 
 % $\Complement{\Fevent_\nummix} \cap \Event_\nummix$ happens, 
% EM algorithm will not converge to global maximum.
Thus, the probability
that the EM algorithm converges to a global maximum is upper bounded
by:
\begin{align*}
\Pb(\Fevent_\nummix) \leq \Pb(\Event_\nummix)\Pb(\Fevent_\nummix\mid\Event_\nummix) +
\Pb(\Complement{\Event_\nummix}) \le \Pb(\Fevent_\nummix \mid \Event_\nummix) +
\Pb(\Complement{\Event_\nummix}) \le e^{-\Omega(\nummix)}.
\end{align*}
To complete the proof for the case when $\nummix = 2^m$ for a positive integer $m$, we need to argue that on the event $\Pb(\Complement{\Fevent_\nummix})$, the log-likelihood of the solution  reached by the EM algorithm can be arbitrarily worse than that of the global maximum. 
We claim that when the event $\Complement{\Fevent_\nummix}$ occurs, the
 EM algorithm returns a solution $\mubold$ for which
 there is at least one urn containing two true
centers which is assigned a single center by the EM algorithm at every iteration 
$t \geq 0$.  As a consequence,
there is at least one true center $\mu_\mixtwo^*$ for which 
we have that 
$\abs{\mu_\mixtwo^*- \mu'_\mixind} \ge \frac{R}{100^m}$ for all
$\mixind=1, \ldots, \nummix$. Now, we claim that we can choose $R$ to be large
enough to ensure an arbitrarily large gap in the likelihood of the EM solution and
the global maximum. Concretely, as $R \to \infty$, we have:
\begin{align*}
\lim_{R \to +\infty} \Like(\mubold) \le \lim_{R \to +\infty}
\frac{1}{\nummix}\E_{X\sim \mathcal{N}(\mu_\mixtwo^*, 1)}
\log\left(\sum_{\mixind=1}^\nummix e^{-\frac{1}{2}(X - \mu_\mixind)^2}
\right)- \frac{1}{\nummix}\log(\nummix\sqrt{2\pi}) =-\infty.
\end{align*}
However, the global maximum $\muboldstar$ has log-likelihood
\begin{align*}
\lim_{R \to +\infty} \Like(\muboldstar) = -\frac{1}{2} - \log(\nummix
\sqrt{2\pi}).
\end{align*}
Once again we can use the continuity of the log-likelihood as a function of $R$ to conclude that
there is a finite sufficiently large $R > 0$ such that the conclusion of Theorem~\ref{thm:main_EM} holds.

%%%%%%%%%%%%%%%%%%%%%%%%%%%%%%%%%%%%%%%%%%

\paragraph{Case $2^{m-1}< \nummix \le 2^m$:}

At a high level, we deal with this case by constructing a
configuration with $2^m$ centers and pruning this down to have
$\nummix$ centers, while ensuring that the resulting urns are still
approximately balanced which in turn ensures that our previous
calculations continue to hold.

Our configuration of true centers in equation~\eqref{hard_problem} can
be viewed as the $2^m$ leaves of a binary tree with depth $\nummix$,
where the vectors $\boldsymbol{\epsilon}$ indexing the true centers
represent the unique path from the root and to the leaf: the value of
$\epsilon_\mixind$ indicates whether to go down to the left child or
to the right child at the $i$-th level of the tree.  We choose
$\nummix$ true centers from the $2^m$ leaves by the following
procedure.  Starting from the root, we assign $\lceil \nummix/2
\rceil$ true centers to the left sub-tree, and assign $\lfloor
\nummix/2 \rfloor$ true centers to the right sub-tree. For any
sub-tree, suppose that it was assigned $l$ true centers, then we
assign $\lceil l/2 \rceil$ true centers to its left subtree and
$\lfloor l/2 \rfloor$ true centers to its right subtree. This
procedure is recursively continued until all the true centers are
assigned to leaves. Each leaf corresponds to a point on the real line
and we choose this point as the location of the corresponding center.

The above construction has the following two properties: first, the
locations of the true centers satisfy the separation requirements
we used in dealing with the case when $\nummix = 2^m$, and further
the assignment of the centers to the left and right urns 
in each case is roughly balanced. 
By leveraging these two properties we can follow essentially the same steps
as we did in the case with $\nummix = 2^m$, and we omit these remaining 
proof details here.

%Finally, we need to show that the log-likelihood of solution the EM
%converge to can be arbitrarily worse than that of the global maximum.
%Suppose that the EM algorithm returns an answer $\mubold$, if it is not a
%global optimum, then there is at least one urn containing two true
%centers which is assigned a single initial center. As a consequence,
%there is at least one true center $\mu_\mixtwo^*$ such that
%$\abs{\mu_\mixtwo^*- \mu'_\mixind} \ge \frac{R}{100^m}$ for all
%$\mixind=1, \cdots, \nummix$. As $R \to \infty$, we have:
%\begin{align*}
%\lim_{R \to +\infty} \Like(\mubold) \le \lim_{R \to +\infty}
%\frac{1}{\nummix}\E_{X\sim \mathcal{N}(\mu_\mixtwo^*, 1)}
%\log\left(\sum_{\mixind=1}^\nummix e^{-\frac{1}{2}(X - \mu_\mixind)^2}
%\right)- \frac{1}{\nummix}\log(\nummix\sqrt{2\pi}) =-\infty
%\end{align*}
%However, for global maximum $\muboldstar$, it has log-likelihood
%$\Like(\muboldstar) \to -\frac{1}{2} - \log(\nummix
%\sqrt{2\pi})$. Since the EM algorithm returns such bad solution $\mubold'$
%with probabilty $1-e^{\Omega(\nummix)}$, by the continuity of
%$\Like(\mubold)$, there is a finite value of $R$ such that the statement
%of Theorem~\ref{thm:main_EM} holds.

%%%%%%%%%%%%%%%%%%%%%%%%%%%%%%%%%%%%%%%%%%%%%%%%%%%%%%%%%%%%%%%%%%%%%%%%%

\subsection{Proof of Theorem~\ref{thm:main_grad_EM}}
\label{SecProofGradEM}

We now embark on the proof of Theorem~\ref{thm:main_grad_EM}. 
The proof follows from a similar outline to the proof of Theorem~\ref{thm:main_EM} 
and we only develop the main ideas here. Concretely, it is easy to verify that in order
to prove the result we only need to establish
the analogue of Lemma~\ref{lem:case_strong}
for the first-order EM algorithm.

Intuitively, we first argue that the first-order EM updates can be viewed as less aggressive 
versions of the corresponding EM updates, and we use this fact to argue that 
Lemma~\ref{lem:case_strong} continues to hold for the first-order EM algorithm.
Concretely, we can compare the update of EM algorithm:
\begin{align*}
\mu_\mixind^{\text{new, EM}} = \frac{\E_{\muboldstar} w_\mixind(\Xrv)
  \cdot \Xrv}{\E_{\muboldstar} w_\mixind(\Xrv)}
\end{align*}
with the update of the first-order EM algorithm:
\begin{align*}
\mu_\mixind^{\text{new, first-order EM}} = \mu_\mixind + \stepsize \E_{\muboldstar} w_\mixind(\Xrv) (\Xrv - \mu_\mixind).
\end{align*}
If for any parameter $\mu_\mixind$, we choose the stepsize $\stepsize
= \frac{1}{\E_{\mu^*} w_\mixind(\Xrv)}$, for the first-order EM
algorithm, then the two updates will match for that parameter.  For
the first-order EM algorithm, we always use a step size $\stepsize \in
(0,1)$, while $\frac{1}{\E_{\muboldstar} w_\mixind(\Xrv)} \ge 1$.
Consequently, there must exist some $\theta_\mixind \in [0, 1]$ such
that
\begin{align*}
\mu_\mixind^{\text{new, first-order EM}} = \theta_\mixind \mu_\mixind
+ (1-\theta_\mixind) \mu_\mixind^{\text{new, EM}}.
\end{align*}
Thus, we see that the first-order EM update is a less aggressive
version of the EM update. An examination of the proof of
Lemma~\ref{lem:case_strong} reveals that this property suffices to
ensure that its guarantees continue to hold for the first-order EM
algorithm, which completes the proof of
Theorem~\ref{thm:main_grad_EM}.

%%%%%%%%%%%%%%%%%%%%%%%%%%%%%%%%%%%%%%%%%%%%%%%%%%%%%%%%%%%%%%%%%%%%%%%%%

\subsection{Proof of Theorem~\ref{prop:escape_saddle}} 
\label{sec:app2}

In this section, we prove Theorem \ref{prop:escape_saddle}. Throughout
this proof, we use the fact that the first-order EM updates with step
size $s \in (0,1)$ take the form
\begin{align}\label{grad_EM_update_new} 
\mubold^{\text{new}} = \mubold + \stepsize \grad \Like(\mubold).
\end{align}
In order to reason about the behavior of the first-order EM algorithm,
we first provide a result that concerns the Hessian of the
log-likelihood.
\begin{lemma}
\label{lem:hess}
For any scalar $s \in (0,1)$ and for any $\boldsymbol{\mu}$, we have
$s \nabla^2 \Like(\boldsymbol{\mu}) \succ -I$.
\end{lemma}
\noindent We prove this claim at the end of the section.  Taking this
lemma as given, we can now prove the theorem's claims. We first show that
the first-order EM algorithm with stepsize $s \in (0,1)$ converges to
a critical point. By a Taylor expansion of the log-likelihood
function, we have
\begin{align*}
\Like(\mubold^{\text{new}}) = & \Like(\mubold) + \langle
\grad\Like(\mubold), \mubold^{\text{new}} - \mubold\rangle +
\frac{1}{2} (\mubold^{\text{new}} - \mubold)^T \grad^2
\Like(\widetilde{\mubold}) (\mubold^{\text{new}} - \mubold),
\end{align*}
for some $\widetilde{\mubold}$ on the line joining $\mubold$ and
$\mubold^{\text{new}}$.  Applying Lemma~\ref{lem:hess} guarantees that
\begin{align*}
\Like(\mubold^{\text{new}}) \geq & \Like(\mubold) + \langle
\grad\Like(\mubold), \mubold^{\text{new}} - \mubold\rangle -
\frac{1}{2s} \|\mubold^{\text{new}} - \mubold\|_2^2.
\end{align*}
From the form~\eqref{grad_EM_update_new} of the gradient EM updates,
we then have
\begin{align*}
\Like(\mubold^{\text{new}}) \geq & \Like(\mubold) + \left(s -
\frac{s}{2} \right) \|\grad\Like(\mubold) \|_2^2.
\end{align*}
Consequently, for any choice of step size $\stepsize \in (0,1)$ and
any point $\mubold$ for which $\grad \Like(\mubold) \neq 0$, applying
the gradient EM update leads to a strict increase in the value of the
population likelihood $\PopLike$.  Since $\PopLike$ is upper bounded
by a constant for a mixture of $\nummix$ spherical Gaussians, we can
conclude that first-order EM must converge to some point. It is easy
to further verify that it must converge to a point for which
$\grad\Like(\mubold) = 0$ which concludes the first part of our proof.

Next we show that the first-order EM algorithm will not converge to
strict saddle points almost surely. We do this via a technique that
has been used in recent
papers~\citep{lee2016gradient,panageas2016gradient}, exploiting
the stable manifold theorem from dynamical systems theory. For this
portion of the proof, it will be convenient to view the first-order EM
updates as a map from the parameter space to the parameter space;
i.e., we define the first-order EM map by:
\begin{align}
\g(\mubold) :=  \mubold + \stepsize \grad \Like(\mubold). \label{gradient_map}
\end{align}
Recalling Definition~\ref{DefnStrict} of strict saddle points, we
denote by $\mathcal{D}_{ss}$ the set of initial points from which the
first-order EM algorithm converges to a strict saddle point.  With
these definitions in place, we can state an intermediate result:

\begin{lemma}[\citep{lee2016gradient,panageas2016gradient}] 
\label{lem:stable_manifold}
If the map $\mubold \mapsto \g(\mubold)$ defined by
equation~\eqref{gradient_map} is a local diffeomorphism for each
$\mubold$, then $\mathcal{D}_{ss}$ has zero Lebesgue measure.
\end{lemma} 

Denote the Jacobian matrix of map $\g$ at point $\mu$ as
$\grad\g(\mubold)$ where $[\grad \g(\mubold)]_{ij} =
\frac{\partial}{\partial \mu_j} g_i(\mubold)$. By
Lemma~\ref{lem:hess}, the Jacobian $\nabla \g(\mubold) = \I +
\stepsize \grad^2 \Like(\mubold)$ is strictly positive definite, and
hence invertible for all $\mubold$, which implies that the map $\g$ is
a local diffeomorphism everywhere.  Furthermore, our random
initialization strategy specifies the distribution of the initial
point $\mubold^{(0)}$ which is absolutely continuous with respect to
Lebesgue measure. Combined these facts with lemma
\ref{lem:stable_manifold}, we have proved Theorem \ref{prop:escape_saddle}.

Finally, the only remaining detail is to prove Lemma~\ref{lem:hess}.
By definition, we have
\begin{align*}
\I + \stepsize \grad^2 \Like(\mubold) &= \underbrace{
\begin{bmatrix}
(1-\stepsize\E w_1(\Xrv)) \I_d & \ldots & 0\\ \ldots & & \ldots \\ 0 &
  \ldots & (1-\stepsize\E w_\nummix(\Xrv)) \I_d
\end{bmatrix}}_{:= \: \DMAT}  + \stepsize \QMAT,
\end{align*}
where the matrix $\QMAT$ has $d$-dimensional blocks of the form
\begin{align*}
\QMAT_{ij} & = \begin{cases}
\E (w_i(\Xrv)-w^2_i(\Xrv)) (\Xrv -\mu_i)(\Xrv -\mu_i)\trans & \mbox{if
  $i = j$} \\
-\E w_i(\Xrv)w_j(\Xrv) (\Xrv - \mu_i)(\Xrv - \mu_j)\trans &
\mbox{otherwise.}
\end{cases}
\end{align*}
Since $w_\mixind(\Xrv) \le 1$ for all $\mixind \in [\nummix]$ and
$\stepsize< 1$, it follows that the diagonal matrix $\Dmat$ is
strictly positive definite.  Consequently, in order to prove Lemma
\ref{lem:hess}, it suffices to show that $\QMAT$ is positive
semidefinite.  Letting $\v = (\v_1\trans, \ldots,
\v_\nummix\trans)\trans$, where $\v_\mixind \in \R^d$, be arbitrary
vectors, we have:
\begin{align*}
\v\trans \QMAT \v= & \sum_{\mixind=1}^\nummix \E w_\mixind(\Xrv)
         [\v_\mixind\trans (\Xrv-\mu_\mixind)]^2
         -\sum_{\mixind=1}^\nummix\sum_{j=1}^\nummix \E
         w_\mixind(\Xrv)w_j(\Xrv) [\v_\mixind\trans
           (\Xrv-\mu_\mixind)][\v_j\trans (\Xrv-\mu_j)] \\
\stackrel{(i)}{\geq} & \sum_{\mixind=1}^\nummix \E w_\mixind(\Xrv)
         [\v_\mixind\trans (\Xrv-\mu_\mixind)]^2
         -\sum_{\mixind=1}^\nummix\sum_{j=1}^\nummix
         \frac{1}{2}\left[\E w_\mixind(\Xrv)w_j(\Xrv)
           [\v_\mixind\trans (\Xrv-\mu_\mixind)]^2 + \E
           w_\mixind(\Xrv)w_j(\Xrv) [\v_j\trans (\Xrv-\mu_j)]^2
           \right] \\
\stackrel{(ii)}{=} & \sum_{\mixind=1}^\nummix \E w_\mixind(\Xrv)
         [\v_\mixind\trans (\Xrv-\mu_\mixind)]^2 -
         \sum_{\mixind=1}^\nummix \E w_\mixind(\Xrv) [\v_\mixind\trans
           (\Xrv-\mu_\mixind)]^2 =0,
\end{align*}
where step (i) uses the elementary inequality $|a b| \leq \frac{1}{2}
(a^2 + b^2)$; and step (ii) uses the fact that $\sum_{i=1}^M w_i(\Xrv)
= 1$ for any $\Xrv$.  This completes the proof.

%%%%%%%%%%%%%%%%%%%%%%%%%%%%%%%%%%%%%%%%%%%%%%%%%%%%%%%%%%%%%%%%%%%%%%%

\subsection*{Acknowledgements}

This work was partially supported by Office of Naval Research MURI
grant DOD-002888, Air Force Office of Scientific Research Grant
AFOSR-FA9550-14-1-001, the Mathematical Data Science program of the
Office of Naval Research under grant number N00014-15-1-2670,
and National Science Foundation Grant CIF-31712-23800.

%%%%%%%%%%%%%%%%%%%%%%%%%%%%%%%%%%%%%%%%%%%%%%%%%%%%%%%%%%%%%%%%%%%%%%%%

\bibliographystyle{plainnat} \bibliography{localEM}

\appendix

\newpage
\section{Proofs of Technical Lemmas} 
\label{sec:app}

The bulk of this section is devoted to the proof of
Lemma~\ref{lem:case_strong}, which is based on a number of technical
lemmas.
%We note that Corollary~\ref{cor:case} is a particular case of
%it.

%%%%%%%%%%%%%%%%%%%%%%%%%%%%%%%%%%%%%%%%%%%%%%%%%%%%%%%%%%%%%%%%%%%%%%%%%%%

\subsection{Proof of Lemma~\ref{lem:case_strong}}
\label{SecProofLemCaseStrong}

Underlying our proof is the following auxiliary result:

\begin{lemma} 
\label{lem:general_calc}
Suppose that:
\begin{enumerate}
\item[(a)] The true distribution is a $\GMM(\muboldstar)$ with
  $\nummix$ components and that all true centers are located in
  $(-\infty, -10a)\cup(a, +\infty)$ with at least one center in $(a,
  3a)$, with $a > \log \nummix + 3$.
\item[(b)] The current configuration of centers has the property that
  for any true center $\mu_\mixtwo^*$ in $(-\infty, -10a)$, there
  exists a current center $\mu_{\mixtwo'}$ such that
  $\abs{\mu_{\mixtwo'}-\mu_\mixtwo^*} \le \abs{\mu_\mixtwo^*}/6$.
\end{enumerate}
Then, for any $\mixind\in [\nummix]$ for which the current parameter
$\mu_\mixind \in [0, 4a]$, we have $\E w_\mixind(X)X \ge 0$.
\end{lemma}
\noindent See Section~\ref{SecProofLemGeneralCalc} for the proof of
this claim.\\

Using Lemma~\ref{lem:general_calc}, let us now prove
Lemma~\ref{lem:case_strong}.  Without loss of generality, we may
assume that \mbox{$\mu_\mixind \in \Ball{c\delta}{2\delta}$,} for some
$\mixind \in [\nummix]$.  Thus, in order to establish the claim, it
suffices to show that after one step of the EM algorithm, the new
iterate $\mu\new_\mixind$ belongs to $\Ball{c\delta}{2\delta}$ as
well.

In order to show that $\mu \new_\mixind \in \Ball{c\delta}{2\delta}$,
note that by the update equation~\eqref{mean_update}, we have
$\mu\new_\mixind = \frac{\E w_\mixind(X)X}{\E w_\mixind(X) }$.  Thus,
it is equivalent to prove that
\begin{align*}
\E w_\mixind(X)(X - (c-2)\delta) \ge 0, \quad\mathrm{and}\quad \E
w_\mixind(X)(X - (c+2)\delta) \le 0.
\end{align*}
The first inequality can be proved by substituting $Z = X -
(c-2)\delta$ and applying Lemma~\ref{lem:general_calc} to $Z$.
Similarly, the second inequality can be proved by defining $Y \defn
(c+2)\delta - X$, and then applying Lemma~\ref{lem:general_calc} to
$Y$.

%%%%%%%%%%%%%%%%%%%%%%%%%%%%%%%%%%%%%%%%%%%%%%%%%%%%%%%%%%%%%%%%%%%%%%%%%%%

\subsection{Proof of Lemma~\ref{lem:general_calc}}
\label{SecProofLemGeneralCalc}

Our proof of this claim hinges on two auxiliary lemmas, which we begin
by stating.  Intuitively, our first lemma shows that if the data are
generated by a single Gaussian, whose mean is at least $\Omega(\log
\nummix)$ to the right of the origin, then it will affect any
$\mu_\mixind\ge 0$, by forcing it to the right no matter where the
other $\{\mu_\mixtwo\}_{\mixtwo\neq \mixind}$ are.

\begin{lemma} 
\label{lem:center_positive}
Suppose that the true distribution is a unit variance Gaussian with
mean $\mu^* \ge a $ for some $a>\log \nummix + 3$, and that the
current configuration of centers, $\mu_1, \cdots, \mu_\nummix$, has
the $i^{\mathrm{th}}$ center $\mu_\mixind\ge 0$. Then we have
\begin{subequations}
\begin{align}
\E w_\mixind(X) X \ge 0.
\end{align}
Furthermore, if $\mu^* \le 3a$, and $0 \leq \mu_\mixind\le 4a$, then:
\begin{align}
\E w_\mixind(X) X \ge \frac{a}{5\nummix} e^{-9a^2/2}.
\end{align}
\end{subequations}
\end{lemma}
\noindent See Section~\ref{SecProofLemCenterPositive} for the proof of
this claim. In a similar vein, if the data is generated by a single
Gaussian far to the left of the origin, and some current center
$\mu_j$ is sufficiently close to it then this Gaussian will force
$\mu_\mixind$ towards the negative direction, but will only have a
small effect on $\mu_\mixind$. More formally, we have the following
result:
\begin{lemma} 
\label{lem:center_negative}
Suppose that the true distribution is a unit variance Gaussian with
mean $\mu^* = -r$, and that the current configuration of centers,
$\mu_1, \cdots, \mu_\nummix$, has the $i^{\mathrm{th}}$ center
$\mu_\mixind\ge 0$ and further has at least one $\mu_j$ such that
$\abs{\mu_j-\mu^*} \le \frac{r}{6}$. Then we have that:
\begin{align}
\E w_\mixind(X)X \ge -3 r e^{-r^2/18}.
\end{align}
\end{lemma}
\noindent See Section~\ref{SecProofLemCenterNegative} for the proof of
this claim. \\

Equipped with these two auxiliary results, we can now prove
Lemma~\ref{lem:general_calc}.  Without loss of generality, suppose
that the centers are sorted in ascending order, and that the
$\ell^{\mathrm{th}}$ true center is the smallest true center in $(0,
+\infty)$. From the assumptions of Lemma~\ref{lem:general_calc}, we
know $\mu_{\ell}^*$ belongs to the interval $(a,3a)$.  Thus, when $X$
is drawn from a Gaussian mixture, we have
\begin{align*}
\E w_\mixind(X)X &= \frac{1}{\nummix}\sum_{j=1}^{\nummix }\E_{X\sim
  \mathcal{N}(\mu_\mixtwo^*, 1)}w_\mixind(X)X \\ &=
\frac{1}{\nummix}\sum_{j=1}^{\ell- 1}\E_{X\sim
  \mathcal{N}(\mu_\mixtwo^*, 1)}w_\mixind(X)X +
\frac{1}{\nummix}\E_{X\sim \mathcal{N}(\mu_{\ell}^*, 1)}w_\mixind(X)X +
\frac{1}{\nummix}\sum_{j=\ell+1}^{\nummix }\E_{X\sim
  \mathcal{N}(\mu_\mixtwo^*, 1)}w_\mixind(X)X.
\end{align*}
We now use Lemma~\ref{lem:center_negative} to bound the first
term. Since $f(y) = -3y\cdot e^{-y^2/18}$ is monotonically increasing
in $[3, +\infty)$, and from the assumptions of
  Lemma~\ref{lem:general_calc}, we have $|\mu^*_j| > -10a > -(9a+2)$
  for all $j < \ell$. Then:
\begin{align*}
\frac{1}{\nummix} \sum_{j=1}^{\ell-1} \E_{X \sim \mathcal{N}
  (\mu_\mixtwo^*, \frac{1}{2})} w_\mixind(X)X \ge -\frac{1}{\nummix}
\sum_{j=1}^{\ell-1}3\mid\mu_\mixind \mid e^{-\mu_\mixind^2/18} \geq
-3(9a+2) e^{-(9a+2)^2/18}.
\end{align*}
By Lemma~\ref{lem:center_positive}, we know that the third term is
non-negative and that the second term can be lower bounded by a
sufficiently large quantity. Putting together the pieces, we find that
\begin{align*}
\E_X w_\mixind(X)X & \geq -3 (9a + 2) e^{-9a^2/2 -2a - \frac{2}{9}} +
\frac{a}{5\nummix ^2} e^{-9a^2/2} \\
& \geq e^{-9a^2/2}\left[ \frac{a}{5 \nummix^2} - 3(9a + 2) e^{-2\log
    \nummix - 6} \right] \\
& \geq \frac{e^{-9a^2/2-6}} {\nummix ^2}\left[80a - 3(9a + 2) \right]
\\
& \geq 0,
\end{align*}
which completes the proof.

%%%%%%%%%%%%%%%%%%%%%%%%%%%%%%%%%%%%%%%%%%%%%%%%%%%%%%%%%%%%%%%%%%%%%%%%%%

\subsection{Proof of Lemma~\ref{lem:center_positive}}
\label{SecProofLemCenterPositive}

Introducing the shorthand $w^* \defn \min_{x \in [1, 2]}
w_\mixind(x)$, we have
\begin{align*}
\E w_\mixind(X) X \ge &\frac{1}{\sqrt{2\pi}}\int_{-\infty}^0 w_\mixind(x)x
e^{-(x-\mu^*)^2/2} \mathrm{d}x + \frac{1}{\sqrt{2\pi}}\int_{1}^2
w_\mixind(x)x e^{-(x-\mu^*)^2/2} \mathrm{d}x \\ &+
\frac{1}{\sqrt{2\pi}}\int_{a}^{3a} w_\mixind(x)x e^{-(x-\mu^*)^2/2}
\mathrm{d}x.
\end{align*}
We calculate the first two terms: for this purpose, the
following lemma is useful:

\begin{lemma} 
\label{lem:wdifference}
For any $\mu_1, \cdots, \mu_\nummix$ where $\mu_\mixind\ge 0$, we have
following:
\begin{align}
\min_{x \in [1, 2]} w_\mixind(x) \ge \frac{1}{\nummix e^2} \max_{x\in (-\infty, 0]}
  w_\mixind(x).
\end{align}
\end{lemma}
\noindent See Section~\ref{SecProofLemWdifference} for the proof of
this claim.
From Lemma~\ref{lem:wdifference}, we have that:
\begin{align*}
&\frac{1}{\sqrt{2\pi}}\int_{-\infty}^0 w_\mixind(x)x e^{-(x-\mu^*)^2/2}
  \mathrm{d}x + \frac{1}{\sqrt{2\pi}}\int_{1}^2 w_\mixind(x)x
  e^{-(x-\mu^*)^2/2} \\ \ge&\frac{1}{\sqrt{2\pi}}
  \left[\int_{-\infty}^0 \nummix e^2w^* x e^{-(x-\mu^*)^2/2}
    \mathrm{d}x + \int_{1}^2 w^* x e^{-(x-\mu^*)^2/2}
    \mathrm{d}x\right] \\ \ge& \frac{w^*}{\sqrt{2\pi}}
  \left[\int_{-\infty}^0 \nummix e^2 (x - \mu^*) e^{-(x-\mu^*)^2/2}
    \mathrm{d}x + \int_{1}^2 x e^{-(x-\mu^*)^2/2}
    \mathrm{d}x\right] \\ \ge& \frac{w^*}{\sqrt{2\pi}} \left[
    -\nummix e^{2-(\mu^*)^2/2} + e^{-(\mu^*-1)^2/2}\right] =
  \frac{w^* e^{-(\mu^*)^2}}{\sqrt{2\pi}} \left[
    e^{\mu^*-1/2}-\nummix e^2\right] \ge 0.
\end{align*}
The last inequality holds since $\mu^* >a > \log \nummix + 3$.  The
third term is always positive, and this finishes the proof of
first claim.

For second claim: if we further know that $\mu^*\le 3a$, and $\mu_\mixind \le
4a$, then for any $x\in [a, 3a]$, $w_\mixind(x) \ge \frac{e^{-9a^2/2}}{\nummix }$,
we have:
\begin{align*}
\frac{1}{\sqrt{2\pi}}\int_{a}^{3a} w_\mixind(x)x e^{-(x-\mu^*)^2/2}
\mathrm{d}x \ge& \frac{1}{\nummix \sqrt{2\pi}} e^{-9a^2/2} a \int_{a}^{3a}
e^{-(x-\mu^*)^2/2} \mathrm{d}x \\
\ge & \frac{a}{\nummix \sqrt{2e\pi}} e^{-9a^2/2} \ge \frac{a}{5\nummix }e^{-9a^2/2}.
\end{align*}
The last inequality is true by integrating over an interval of length 1
around $\mu^*$ contained in $(a, 3a)$.

%%%%%%%%%%%%%%%%%%%%%%%%%%%%%%%%%%%%%%%%%%%%%%%%%%%%%%%%%%%%%%%%%%%%%%%%%%%%

\subsubsection{Proof of Lemma~\ref{lem:wdifference}}
\label{SecProofLemWdifference}

We split the proof into two cases.

\paragraph{Case $\mu_\mixind \in [0, 2]$:}
In this case, we are guaranteed that $\max_{x\in (-\infty, 0]}w_\mixind(x)
  \le 1$.  Also, for any $x\in [1, 2]$, we have:
\begin{align}
w_\mixind(x) = \frac{e^{-(x-\mu_\mixind)^2/2}}{\sum_j e^{-(x-\mu_j)^2/2}}\ge
\frac{1}{\nummix e^2},
\end{align}
which proves the required result.

\paragraph{Case $\mu_\mixind > 2$:}  In this case, we have
\begin{align*}
w_\mixind(x) = & \frac{e^{-(x-\mu_\mixind)^2/2}}{\sum_j e^{-(x-\mu_j)^2/2}} =
\frac{1}{\sum_{j\neq i} \frac{1}{\nummix -1} + e^{[(x-\mu_\mixind)^2 -
      (x-\mu_j)^2]/2}}\\ = & \frac{1}{\sum_{j\neq i} \frac{1}{\nummix -1} +
  e^{(\mu_\mixind - \mu_j)(\mu_\mixind + \mu_j - 2x)/2}} = \frac{1}{\sum_{j\neq i}
  A_{ij}(x)},
\end{align*}
where $A_{ij}(x) \defeq \frac{1}{\nummix -1} + e^{(\mu_\mixind - \mu_j)(\mu_\mixind +
  \mu_j - 2x)/2}$. It suffices to show that
\begin{align}
\label{EqnSufficient}
A_{ij}(x) & \leq \nummix A_{ij}(x') \quad \mbox{for any $x\in[1,2]$, $x' \in
  (-\infty,0]$ and $j \in [\nummix ]$.}
\end{align}
Using this, we know:
\begin{align}
w_\mixind(x) = \frac{1}{\sum_{j \neq i} A_{ij}(x)} \ge \frac{1}{ \sum_{j\neq
    i} \nummix A_{ij}(x')} = \frac{1}{\nummix} w_\mixind(x'),
\end{align}
and the claim of Lemma~\ref{lem:wdifference} easily follows.  In order
to establish the claim of equation~\eqref{EqnSufficient}, we note that
if $\mu_j \le \mu_\mixind$, then since $x' < x$, we have
\begin{align*}
(\mu_\mixind - \mu_j)(\mu_\mixind + \mu_j - 2x) \le (\mu_\mixind - \mu_j)(\mu_\mixind + \mu_j
  - 2x'),
\end{align*}
which implies that $A_{ij}(x) \le A_{ij}(x')$.
If $\mu_\mixind < \mu_j$, then we know:
\begin{align}
(\mu_\mixind - \mu_j)(\mu_\mixind + \mu_j - 2x) <0.
\end{align}
This implies $A_{ij}(x) \le \frac{1}{\nummix -1} + 1 = \frac{\nummix }{\nummix -1}$. On the
other hand, we always have $A_{ij}(x') \ge \frac{1}{\nummix -1}$, this gives
$A_{ij}(x) \le \nummix  A_{ij}(x')$, which finishes the proof.

%%%%%%%%%%%%%%%%%%%%%%%%%%%%%%%%%%%%%%%%%%%%%%%%%%%%%%%%%%%%%%%%%%%%%%%

\subsection{Proof of Lemma~\ref{lem:center_negative}}
\label{SecProofLemCenterNegative}

We have 
\begin{align*}
\E w_\mixind(X)X \ge \frac{1}{\sqrt{2\pi}}\int_{-\infty}^{-2r/3}
w_\mixind(x)x e^{-(x-\mu^*)^2/2} \mathrm{d}x +
\frac{1}{\sqrt{2\pi}}\int_{-2r/3}^0 w_\mixind(x)x e^{-(x-\mu^*)^2/2}
\mathrm{d}x.
\end{align*}
For the first term, we know for any $x \in (-\infty,-2r/3]$, we have:
\begin{align*}
w_\mixind(x) \le \frac{e^{-x^2/2}}{e^{-(x-\mu_j)^2/2}} = e^{-x\mu_j +
  \mu_j^2/2} \le e^{-\frac{2r}{3}\mu_j + \mu_j^2/2} \le
e^{-\frac{7}{72}r^2} \le e^{-r^2/18}.
\end{align*}
The second last inequality is true since $\mu_j\ge
-\frac{7r}{6}$. Thus, we know:
\begin{align*}
\frac{1}{\sqrt{2\pi}}\int_{-\infty}^{-2r/3} w_\mixind(x)x
  e^{-(x-\mu^*)^2/2} \mathrm{d}x &\ge
  \frac{e^{-r^2/18}}{\sqrt{2\pi}}\int_{-\infty}^{-2r/3}x
  e^{-(x-\mu^*)^2/2} \mathrm{d}x \\ \ge&
  \frac{e^{-r^2/18}}{\sqrt{2\pi}}\left[\int_{-\infty}^{-2r/3}
    (x-\mu^*) e^{-(x-\mu^*)^2/2} \mathrm{d}x +
    \mu^*\sqrt{2\pi} \right] \\ \ge&
  \frac{e^{-r^2/18}}{\sqrt{2\pi}}\left[-\frac{1}{2}e^{-r^2/18}
    -r\sqrt{2\pi}\right] \ge -2re^{-r^2/18}.
\end{align*}
For the second term, we have:
\begin{align*}
\frac{1}{\sqrt{2\pi}}\int_{-2r/3}^0 w_\mixind(x)x e^{-(x-\mu^*)^2/2}
\mathrm{d}x \ge& -\frac{2r}{3\sqrt{2\pi}}
\int_{-2r/3}^0e^{-(x-\mu^*)^2/2} \mathrm{d}x
\\ \ge&-\frac{2r}{3\sqrt{2\pi}} \int_{-2r/3}^{+\infty}
e^{-(x-\mu^*)^2/2} \mathrm{d}x \ge -\frac{2r}{3} e^{-r^2/18}.
\end{align*}
Putting the pieces together we obtain,
\begin{align*}
\E w_\mixind(X)X \ge -(\frac{2}{3}+2) e^{-r^2/18} \ge -3r e^{-r^2/18},
\end{align*}
as desired.

\end{document}